\documentclass[11pt]{article}

\usepackage{geometry}
\usepackage[table, dvipsnames]{xcolor}
\usepackage{setspace}
\usepackage{caption} 
\usepackage{graphicx}
\usepackage{url}
\usepackage{longtable}
\usepackage[framemethod=tikz]{mdframed}
\usepackage{enumitem}
\usepackage{multirow}

\usepackage{mathptmx}
\usepackage[T1]{fontenc}

\usepackage{natbib}
\setcitestyle{aysep={}}

\newcounter{tfnumquote}
\newenvironment{tfquote}{%
\refstepcounter{tfnumquote}%
\quote}{\unskip~(Text Fragment TF\thetfnumquote)\endquote}

\newcounter{tnumquote}

\usepackage{graphicx}
\usepackage{todonotes}
\usepackage{url}
\usepackage{float}
\usepackage{comment}
\usepackage[framemethod=tikz]{mdframed}
\usepackage{xcolor, soul}
\sethlcolor{cyan}
\usepackage{soul}

\RequirePackage[bookmarks,
                urlcolor=black,
                citecolor=black,
                linkcolor=black,
                colorlinks,
                hyperfigures]{hyperref}

\newtheorem{concept}{Concept}

\definecolor{darkgreen}{rgb}{.0,0.53,.0}
\newcommand\srm[1]{\textcolor{black}{#1}}
\newcommand\jvb[1]{\textcolor{black}{#1}}
\newcommand\nk[1]{\textcolor{black}{#1}}

\title{Conversational Process Modeling: Can Generative AI Empower Domain Experts in Creating and Redesigning Process Models? } 

\author{Nataliia Klievtsova$^{1}$, Janik-Vasily Benzin$^{1}$, Timotheus Kampik$^{2}$, \\Juergen Mangler$^{1}$, and Stefanie Rinderle-Ma$^{1}$\\
$^{1}$Technical University of Munich, Germany; \\TUM School of Computation, Information and Technology, \\
\{nataliia.klievtsova, janik.benzin, juergen.mangler, stefanie.rinderle-ma\}@tum.de\\
$^{2}$SAP Signavio, Berlin, Germany, timotheus.kampik@sap.com}

\date{}

\begin{document}

\maketitle

\begin{abstract}
  AI-driven chatbots such as ChatGPT have caused a tremendous hype lately. For BPM applications, several applications for AI-driven chatbots have been identified to be promising to generate business value, including explanation of process mining outcomes and preparation of input data. However, a systematic analysis of chatbots for their support of conversational process modeling as a process-oriented capability is missing. This work aims at closing this gap by providing a systematic analysis of existing chatbots. Application scenarios are identified along the process life cycle. Then a systematic literature review on conversational process modeling is performed, resulting in a taxonomy of application scenarios for conversational process modeling, including paraphrasing and improvement of process descriptions. In addition, this work suggests and applies an evaluation method for the output of AI-driven chatbots with respect to completeness and correctness of the process models. This method consists of a set of KPIs on a test set, a set of prompts for task and control flow extraction, as well as a survey with users. Based on the literature and the evaluation, recommendations for the usage (practical implications) and further development (research directions) of conversational process modeling are derived.

\end{abstract}


\section{Introduction}
\label{ref:intro}

AI-powered chatbots \textsl{``have a considerable impact in many domains directly related to the design, operation, and application of information systems''} and at the same time need to be handled with care \citep{teub2023}, as they provide their users with information without considering their own technology's limitations. Business process management as an information systems discipline seems a viable candidate to benefit from chatbots and and hence from the recent advances in large language models, in particular, when supporting users in creating and improving process-related content, most prominently process models and process descriptions. 

Process models enable participants to understand the processes in which they are involved~\citep{bpmn_value} and to improve business performance~\citep{bpmn_chal}. 
However, errors in process models may have adverse business consequences~\citep{profit}, and may lead to problems during process execution and quality issues in the process outcome~\citep{anti}.

Currently the creation of process models is often based on the interaction between domain experts having the knowledge of the process and process modellers/analysts capable of process modelling and analysis techniques. 
Hence, the acquisition of as-is models can consume up to $60$\% of the time spent on process management projects according to~\citep{bpmngen}. \srm{This manual effort might be reduced by discovering process models automatically from event data by applying process mining techniques. However, it still requires the existence of such data as well as data preparation effort.} 

The overarching question of this work is thus how and to which degree chatbots can augment or even replace the process modeller/analyst when creating process models through conversational modelling with the domain expert.
Conversational modelling means conversation flow modelling where the chatbot can receive and interpret inputs from the user (i.e., follow-up questions, unexpected inputs, or changes of topic) and provide appropriate responses that keep the conversation coherent~\citep{CM}.

This \nk{overarching} question can be broken down into the following research questions:

\noindent\textbf{RQ1} How can CM methods/tools be employed for process modelling?

 \noindent\textbf{RQ2} Which CM methods/tools exist for process modelling?

\noindent\textbf{RQ3} How can we evaluate CM methods/tools with respect to process modelling? 

\noindent\textbf{RQ4} Which implications do Chatbots have for BPM modelling practice/research?

RQ1 -- RQ4 are tackled \srm{along a well-defined research process presented in Sect. \ref{sec:researchmethod} that comprises the following steps\footnote{\srm{Note that this work constitutes an extension of \citep{DBLP:conf/bpm/KlievtsovaBKMR23} by providing a LLM-based task extraction method, a qualitative evaluation of process model correctness, and a user survey, along with the extended findings and implications.}}. At first,the concept of conversational modelling is transferred to conversational process modelling. Based on the concept of conversational process modelling, initial application scenarios are derived in accordance with} the process life cycle (cf. Sect. \ref{sec:convermod}). These initial application scenarios provide the keywords for the subsequent literature review (cf. Sect. \ref{sec:sota}) which aims at refining the scenarios along a taxonomy of existing approaches. 
\srm{Section \ref{sec:per} shows how LLMs such as GPT3 and GPT4 can be employed for task and control flow extraction, in connection with the graphical output of the process model using Mermaid.js and Graphviz. The resulting process models are evaluated in a quantitative way w.r.t. their completeness, using a set of key performance indicators, and in a qualitative way regarding their correctness. Moreover, a survey with users is conducted in order to compare gold standard process models with LLM-generated ones regarding completeness and correctness. The findings are intriguing, as users prefer the LLM-based models independent of their experience in process modelling. The key performance indicators are elaborated based on a test set from higher education and the evaluation is conducted based on a training set from literature. The evaluation results are consolidated and yield a set of practical implications and research directions presented in Sect. \ref{ref:vision}. }

\section{Research Method}
\label{sec:researchmethod}

\srm{This study follows the design science research (DSR) methodology structured and fleshed out according to the following core dimensions of DSR projects as presented in \citep{DBLP:journals/electronicmarkets/BrockeM19}:}

\begin{figure}[htb!]
    \centering
    \includegraphics[width=\textwidth]{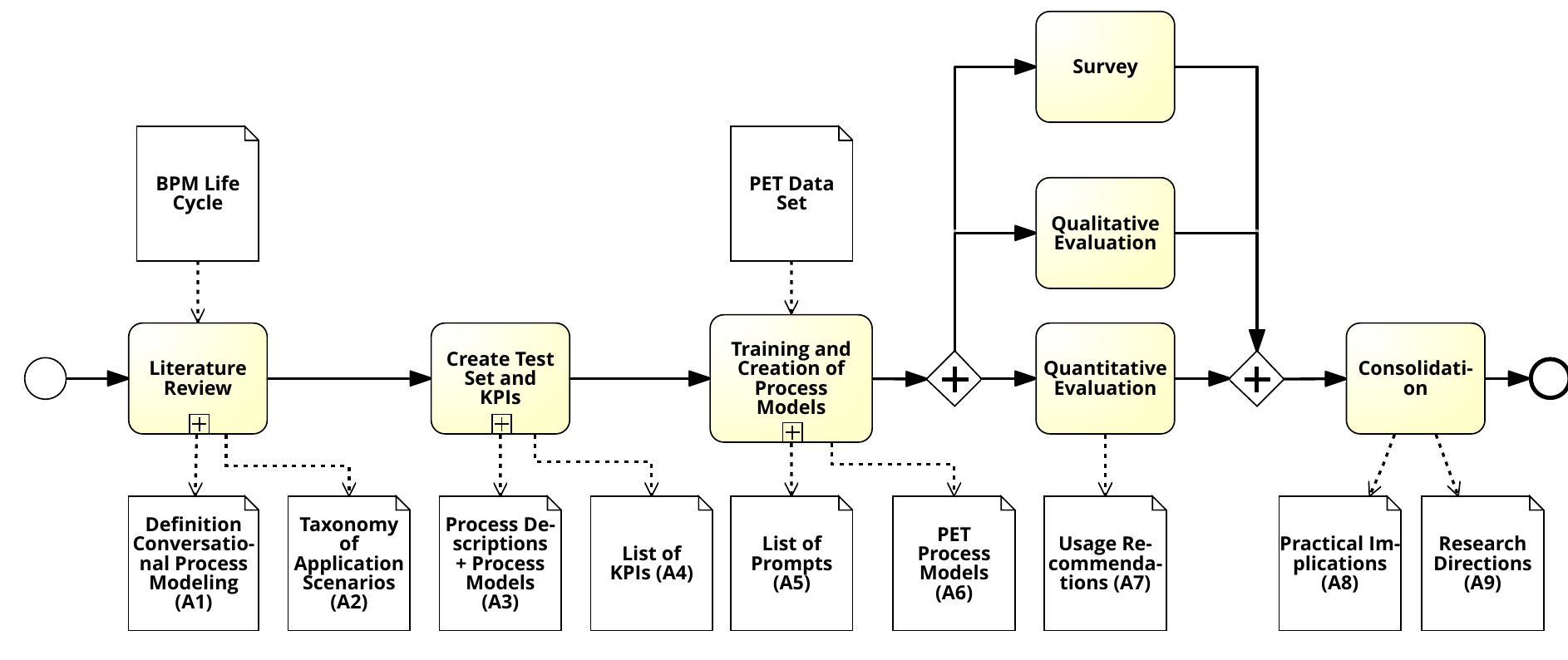}
    \caption{Abstracted Research Method (Modelled in BPMN)}
    \label{fig:convermodprocessmethodabstract}
\end{figure}

\srm{
\begin{itemize}
    \item \textsl{Problem description:} Business process modelling can be cumbersome, complex, and expensive, in particular, when process modellers and domain experts do not efficiently communicate~\citep{bpmngen}. Hence, the question is whether domain experts can be empowered by using LLMs and chatbots and as a consequence, business process modelling projects can be streamlined and simplified. 
 \item\textsl{Input knowledge:} As input, we use the business process life cycle (see, e.g., \citep{dumas_fundamentals_2013}) and the PET dataset as provided in \citep{bpmnentity} as training dataset. This is reflected as input data objects in the research process model depicted in Fig. \ref{fig:convermodprocessmethodabstract}. 
 \item\textsl{Research process:} The main tasks of the research process are depicted in Fig. \ref{fig:convermodprocessmethodabstract}. The process starts with a \textsl{Literature Review}. It consists of an analysis of the BPM life cycle in order to determine in which life cycle phases conversational process modelling can be applied. This leads to a definition of conversational process modelling (Artefact 1, A1) and yields the keywords for the subsequent systematic literature review. The systematic literature review, in turn, results in a taxonomy of application scenarios for conversational process modelling (Artefact 2, A2). In the second phase \textsl{Create Test Set and KPIs}, we create a set of process descriptions with the associated process models and quality assessments of the models (Artefact 3, A3). \nk{Subsequently}, a list of key performance indicators (KPIs) is elicited in order to enable the quantitative assessment of the process models in the sequel (Artefact 4, A4). This is followed by \nk{the} task \textsl{Training and Creation of Process Models}. Here, the PET dataset \citep{bpmnentity} is used for training, especially for engineering the prompts, resulting in a list of prompts (Artefact 5, A5) and the PET process models (Artefact 6, A6). This dataset is then evaluated in parallel in a \textsl{Quantitative Evaluation}, a \textsl{Qualitative Evaluation}, and a \textsl{Survey}. The quantitative evaluation utilised the PET process models and the KPIs created based on the test dataset, resulting a selection of prompts which is then evaluated during a control evaluation. 
 The quantitative evaluation results in usage recommendations, e.g., concerning the prompts (Artefact 7, A7). \srm{The qualitative evaluation assesses the correctness of the produced process models. The survey tests whether users prefer the process models created as ``gold standard'' over the process models created by the LLMs. These extensive evaluations yield practical implications (Artefact 8, A8) and research directions (Artefact 9, A9). }
  \item\textsl{Key concepts:} The key concept of this study is conversational process modelling, i.e., the interactive creation and redesign of process models from process descriptions based on an interactive conversation between domain experts and chatbots. The goal is to investigate the quality of the resulting process models for state of the art generative AI approaches in a quantitative and qualitative way. 
 \item\textsl{Solution description:} The solution of the addressed problem is based on the conversational process modelling process comprising of a list of prompts, a training and test dataset, and a set of KPIs to measure the quality of the created and/or redesigned process models. 
 \item\textsl{Output knowledge:} Artefacts 1--8 as reflected by the 
 output data objects of the research process shown in Fig. \ref{fig:convermodprocessmethodabstract} constitute the output knowledge, i.e., a definition of the concept of conversational process modelling, a taxonomy of application scenarios, process descriptions and process models based on a test dataset, a list of KPIs to measure the output quality, a list of prompts, the training PET process models, usage recommendations, as well as research directions. 
 \end{itemize} }

\section{Conversational Process Modelling}
\label{sec:convermod}

Only few papers address conversational modelling, mostly by focusing on the design of virtual human agents (aka chatbots), e.g., \citep{DBLP:conf/iva/RossenLL09,CM}. However, there is no common understanding of conversational \textbf{process} modelling yet and we hence provide informal Concept~\ref{concept:convermod} which takes up characteristics of conversational modelling regarding the participants in the conversation, i.e., the domain expert and the chatbot, and the iterative nature of the conversation.

\begin{concept}[Conversational Process modelling \srm{(ConverMod)}] describes the process of creating and improving process models and process descriptions based on the iterative exchange of questions/answers between domain experts and chatbots.
\label{concept:convermod}
\end{concept}

\begin{figure}[htb!]
    \centering
    \includegraphics[width=\textwidth]{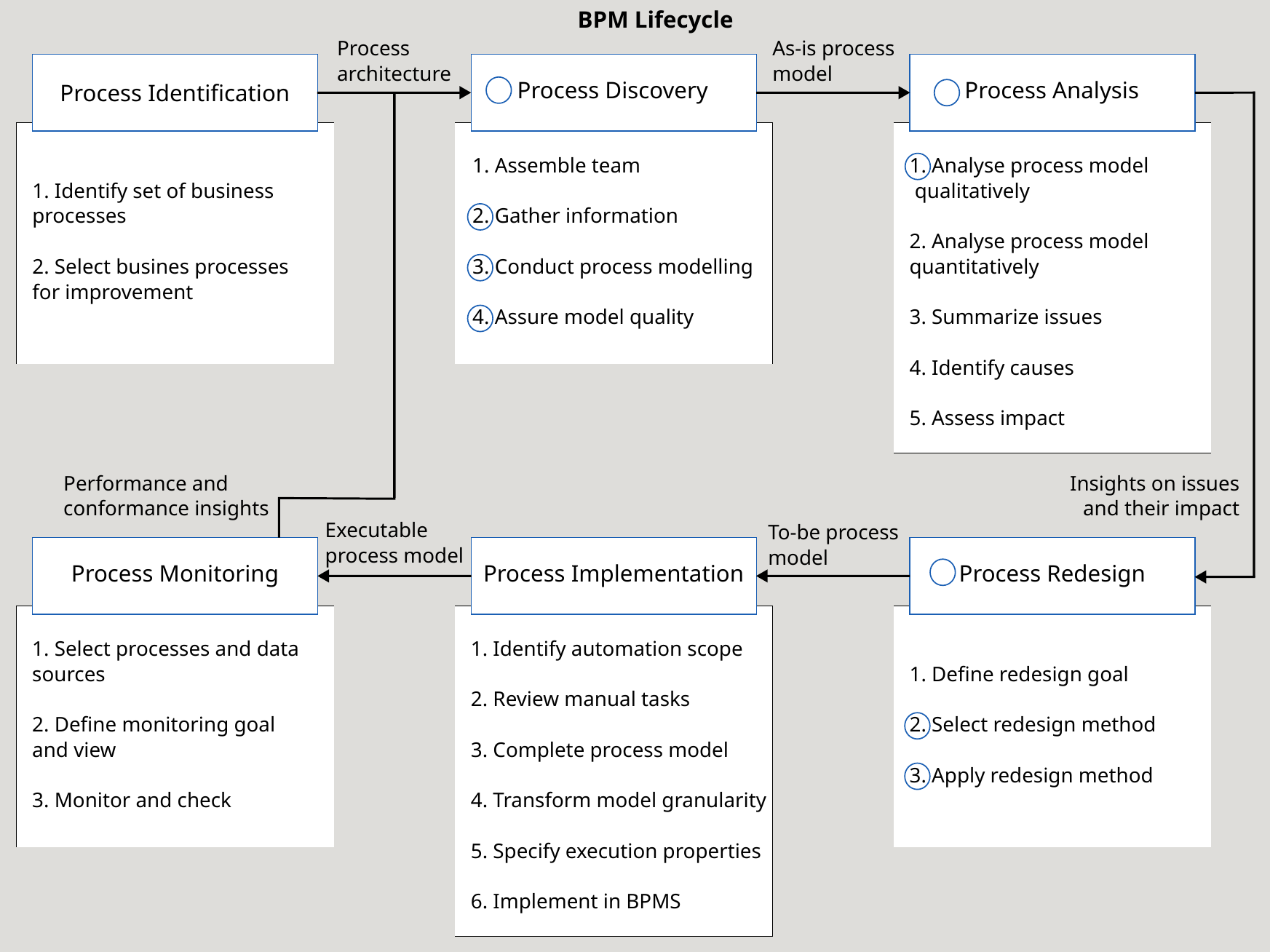}
    \caption{BPM Life Cycle Phases with ConverMod Applications}
    \label{fig:bpmlifecycle}
\end{figure}

Concept \ref{concept:convermod} reflects the overarching goal of ConverMod, i.e., to enable process modelling and improvement based on interaction between the domain expert and the chatbot, instead of interaction between the domain expert and the process analyst/modeller. \srm{In order to understand in which process modelling scenarios ConverMod can be applied, we start with analysing the BPM life cycle as provided by \citep{dumas_fundamentals_2013} (cf. Fig. \ref{fig:bpmlifecycle}). The BPM life cycle is chosen as it provides a systematic structuring of the different process-oriented tasks and capabilities towards creating business value. We assume} that that ConverMod is exclusively based on domain expert/chatbot interaction and does not employ any other tool. In the conclusion, we sketch how ConverMod can be extended if chatbot usage is augmented by other tools such as process simulators.

\textbf{Process discovery} subsumes a range of methods for creating process models (and is not to be confused with process discovery in process mining, which is based on event logs). The typical input in a process discovery project consists of textual process descriptions gathered based on interviews or workshops. Based on the process descriptions, process models are created by process modellers/analysts. We identified the following steps as suitable for being supported by chatbots: (1) gathering the process descriptions for creating the process model. This also includes the preparation of the process descriptions, i.e., to increase the quality of the process description in terms of, for example, precision, e.g. through automatic paraphrasing.
(2) taking a process description as an input and producing a process model (accompanied by the process description). Here, the chatbot can be employed for analysing the text and extracting process model relevant information such as activities and their relations as well as actors \citep{bpmnentity}.
Finally (3) assessing a process model (with the accompanying process description), regarding  model quality based on quality metrics such as cohesion~\citep{DBLP:journals/cii/VanderfeestenRA08} and \nk{broader} guidelines such as number of elements or label style \citep{DBLP:journals/bpmj/AvilaSMT21}.

The \textbf{process analysis} phase builds the bridge between the as-is process model created in the process discovery phase and the to-be model created in the process redesign phase. It is concerned with the qualitative and quantitative assessment of a process model. A qualitative analysis comprises, for example, an assessment whether or not certain activities can be automated; this can then be analogously reflected by an action recommendation, e.g., if the automation potential is not fully exploited, yet. The chatbot can support this assessment based on the extracted activities in the process discovery phase. The results of the qualitative assessment can then be used in the process redesign phase for corresponding redesign actions.
Quantitative process analysis comprises, for example, the detection of bottlenecks based on process simulations. As mentioned before, for this work, we assume that the chatbot is used without invoking further tools and systems such as a process simulator. Hence, quantitative process analysis does not include tasks for ConverMod at this stage, but for future work (cf. Sect. \ref{sec:disc}).

\textbf{Process redesign} comprises the definition of the redesign goal which again is considered a managerial task. The chatbot can support the domain expert by proposing existing redesign methods such as Lean Six Sigma, as well as in querying models (see~\citep{DBLP:books/sp/22/P2022}) or applying redesign instructions. Especially important is refactoring of process descriptions, based on existing guidelines on process model refactoring or catalogues of process smells such as the one presented by \citep{DBLP:journals/cii/WeberRMR11}.

The phases of \textbf{process implementation} and \textbf{process monitoring} are considered as a part of future work of ConverMod as they will require the invocation of additional tools and systems such as a process engine or process-aware information system.

Table \ref{tab:applife} summarises the initial application scenarios for ConverMod along the process life cycle phases and steps which constitute the input for the subsequent literature and test set based analyses as well as the literature survey. 

\begin{table}[htb!]
    \centering
    \caption{Application Scenarios and Chatbot Tasks along BPM Life Cycle}
          \begin{tabular}{p{4cm}|p{3.6cm}|p{3.2cm}|p{2.5cm}}
         \textbf{\# application} &  \textbf{input} & \textbf{output} & \textbf{chatbot task} \\\hline
         1. gather information & process description & process description & paraphrase \\\hline
         2. process modelling &  process description & process model, process description & extract\\\hline
         3. assure model quality & process model, process description, process modelling guidelines and metrics & quality issues, refined process model, refined process description& compare and assess\\\hline
         4. select redesign method & collection of process models and process descriptions & redesign method, selection of process models and process descriptions & select method, query models \\\hline
         5. apply redesign method & collection of process models and process descriptions, redesign method & collection of process models and process descriptions & query and refactor models\\
    \end{tabular}
    \label{tab:applife}
\end{table}

\begin{figure}[ht]
    \centering
    \includegraphics[width=\textwidth]{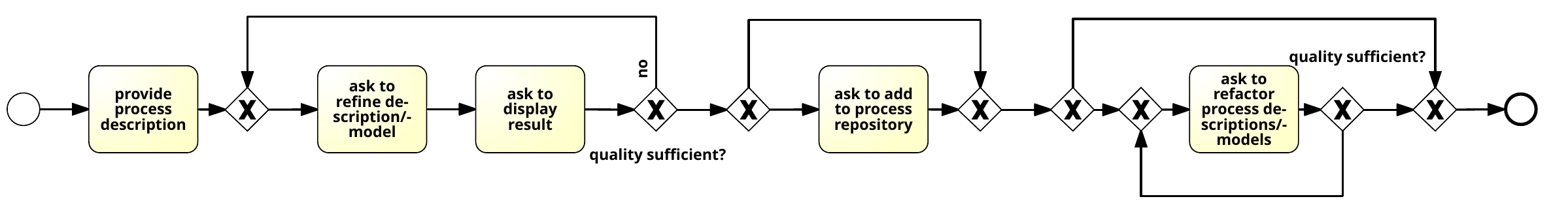}
    \caption{The Process of ConverMod (modeled in BPMN)}
    \label{fig:convermodprocess}
\end{figure}

The BPMN model depicted in Fig. \ref{fig:convermodprocess} assembles and refines the application scenarios, together with their input, outputs, and related chatbot tasks as summarised in Table \ref{tab:applife} into a generic process model for ConverMod, reflecting its interactive and iterative characteristics: at first, the domain expert provides a process description which is refined ($\rightarrow$ paraphrase) and the results are displayed ($\rightarrow$ extract). Then an assessment of the result quality is conducted ($\rightarrow$ compare and assess). If the quality is insufficient, the process models/descriptions are refined ($\rightarrow$ query, refactor), possibly based on a specific method ($\rightarrow$ select method), until the quality reaches a sufficient level.

\srm{Section \ref{sec:convermod} provides Concept \ref{concept:convermod} as a definition of ConverMod and an initial taxonomy of application scenarios which correspond to Artefact 1 and 2 in the research process depicted in Fig. \ref{fig:convermodprocessmethodabstract}. The latter serves as input for the subsequent literature analysis in Sect. \ref{sec:sota}.}

\section{State of the Art}
\label{sec:sota}
\sethlcolor{cyan}

The literature analysis consists of two steps, i.e., i) a pre-review based on the initial application scenarios and life cycle phases summarised in Table \ref{tab:applife} and based on the outcome of the pre-review, ii) a more generalised review including, for example, NLP-based methods for the extraction of model information from process descriptions. i) and ii) follow the guiding principles of \citep{kitchenham_procedures_2004}.

\noindent\textbf{i) Pre-review:}
The pre-review is conducted based on the keywords resulting from
 building the cross product of the application scenarios and the keyword \texttt{chatbot}, which is summarized in Table \ref{tab:applife}, e.g,  \texttt{``process modeling'' chatbot}. These keywords are then used in the title search (\texttt{allintitle}) on \url{google.scholar.com}\footnote{last accessed 2023-10-25}.
Next, we use the keywords resulting from the cross product of the application scenarios and chatbot task, e.g., \texttt{``process modeling'' paraphrase}
and the keywords resulting from the cross product of the keyword \texttt{conversational} and the application scenarios (allintitle), e.g., \texttt{conversational ``process modeling''}.
In order to broaden the pre-review, we repeated the search for application and chatbot, but without the keyword \texttt{process}. 
Most of these searches result in $0$ or very few hits, which were rejected due to quality issues or domain irrelevance. 

The pre-review did not yield deeper insights into techniques, opportunities, and limitations of ConverMod. The results rather point towards generalising the keywords used for the search, particularly covering methods based on Natural Language Processing (NL). Hence, for the \textbf{ii) second search}, we used  \url{scholar.google.com} to produce Table~\ref{tab:slr}. It shows a list of 59 papers relevant for a wide variety of related topics. Papers were selected based on the occurrence of the enumerated keywords (Selection Criteria) in the abstract or the title (for the first 20 hits).

In the following, we discuss the literature collected in Table~\ref{tab:slr} regarding six fundamental questions that partly correspond to the research questions and partly to the pointers derived from the pre-review.

\noindent\textbf{How do chatbots work, and what are important areas of application?} A chatbot is a type of a computer program simulating human-computer interaction (i.e., conversations) to solve particular user problems~\citep{chatbot}. Chatbots work by processing language input from humans (furthermore referred to as NLP~\citep{nlp,meurers2012natural}), and reacting to it. The interpretation of human input is achieved through a set of rules~\citep{lim2020experimental,dihyat2021can,choaresearch}, or by utilising large language models (LLMs)~\citep{liu2021ai}, which are trained to understand the input's meaning/intent/context~\citep{types,canizares2022automating} and generate new content based on different statistical and probabilistic techniques. According to~\citep{applications} the main areas of chatbot applications are human resources, e-commerce, learning management systems, customer service, and sales.

\begin{longtable}{|p{5cm}|l|p{3.5cm}|l|p{3.7cm}|}
    \caption{Literature Queries, Hits, and Selections}
    \label{tab:slr}\\
         \hline
         \textbf{Query (allintitle:)} & \textbf{Hits} & \textbf{Selection Criteria} & \textbf{\#} & \textbf{List} \\
         \hline
         chatbot technology overview  & 1 &  & 1 & \citep{chatbot} \\
         \hline
         Natural language processing  & 10400 & automated NLP & 2 &  \citep{nlp},\citep{meurers2012natural}\\
         \hline
         nlp Chatbot Development & 7 & deep learning & 1 & \citep{areas} \\
         \hline
         chatbots business processes  & 2 & capability to learn & 1 & \citep{learnmodel} \\
         \hline
         Chatbot integration & 32 & chatbot integration & 1 & \citep{survey} \\
         \hline
         quark chatbot & 1 &  & 1 & \citep{quark} \\
         \hline
         ((Chatbots) OR (chatbot)) Process Models  & 2 & process model & 1 & \citep{modchat} \\
         \hline
         reasoning processes descriptions & 3 &  & 1 & \citep{atdp} \\
         \hline
         process model generation" & 15 & text & 1 & \citep{bpmngen} \\
         \hline
         generating BPMN  diagram  & 2 & text & 1 & \citep{bpmngen3} \\
         \hline
         business process (model) OR (models) generating & 34 & Natural Language,document sources & 2 & \citep{9946460},\citep{bpmngen4}\\
         \hline
         extracting business process language models & 2 & NLP, language model & 2 & \citep{8010715},\citep{bpmnentity} \\
         \hline
         AI based language models & 2 & NLP, LMs & 1 & \citep{liu2021ai} \\
         \hline
         large language models & 628 & NLP, BPMN & 3 & \citep{min2021recent},\citep{witteveen2019paraphrasing} ,\citep{kojima2022large}\\
         \hline
         BPMN generation & 22 & NLP, LMs & 1 & \citep{maqbool2019comprehensive} \\
         \hline
         ``process extraction'' from text & 6 & text, textual information & 1 & \citep{bpmngen2},\citep{bellan2021process},\citep{bellan2022process} \\
         \hline
         ``knowledge graphs'' chatbots  & 5 & NLP, LMs & 1 & \citep{avila2020conquest},\citep{graph},\citep{graph2},\citep{omar2023chatgpt},\newline\citep{patsoulis2021integration} \\
         \hline
         chatbots BPMN modeling & 0 & --- & --- & --- \\
         \hline
         chatbots graph generation & 0 & --- & --- & --- \\
         \hline
         ((model based) OR (model-based)) & 12 & NLP, BPMN, UML & 1 & \citep{uml} \\
         \hline
         generate graphs chatbots & 0 & --- & --- & --- \\
         \hline
         generate graphs plain text & 0 & --- & --- & --- \\
         \hline
         BPMN modeling chatbots & 0 & --- & --- & --- \\
         \hline
         low-code chatbot development & 1 &  & 1 & \citep{low-code} \\
         \hline
         generating texts models  & 2 & process model & 1 & \citep{modtext} \\
         \hline
         declarative process model generation & 0 & --- & --- & --- \\
         \hline
         process models chatbot & 1 & --- & 1 & \citep{dpm} \\
         \hline
         process conversational agents & 7 & BPMN & 2 & \citep{pacas}, \citep{model} \\
         \hline
         rule based chatbots & 5 & natural language, AIML & 3 & \citep{lim2020experimental}, \citep{dihyat2021can}, \citep{choaresearch} \\
         \hline
         chatbot designs & 4 & natural language  & 2 & \citep{types}, \citep{canizares2022automating}\\
         \hline
         Process Models Chatbots  & 1 & --- & 1 & \citep{modchat} \\
         \hline
         mining models from text & 11 & process model  & 1 & \citep{Riefer2016}\\
         \hline
         automatic generation bpmn & 5 & from BPMN, process model  & 3 & \citep{etl},\citep{moise},\citep{checklist}\\
         \hline
         text information extraction & 539 &  unstructured text, semi-structured text & 7 & \citep{jiang2012information},\citep{soderland1999learning},\citep{mooney2005mining},\citep{rau1989information},\newline\citep{tang2013hybrid},\citep{ciravegna2001adaptive},\citep{rahman2022characterizing}\\
         \hline
         text data augmentation methods & 8 & methodology & 1 & \citep{methodology} \\
         \hline
         data augmentation approaches nlp & 1 &  & 1 & \citep{augsur}\\
         \hline
         easy data augmentation techniques & 4 & data augmentation & 3 & \citep{eda1},\citep{eda2},\citep{eda3} \\
         \hline
         automatic machine translation paraphrasing & 3 & paraphrasing & 2 & \citep{mt1},\citep{yoshimura2019filtering}\\
         \hline
         paraphrasing automatic evaluation & 7 &  & 2 & \citep{kauchak2006paraphrasing},\citep{bleu1},\citep{yoshimura2019filtering} \\
         \hline
         llm for bpmn & 242 & BPM, BPM life cycle, process modelling, BPMN  & 6 & 
         \citep{VidgofBM23},\citep{Jana},\citep{GPTMining},\citep{LPM},\citep{JsonAbs},\citep{DBLP:conf/bpm/KlievtsovaBKMR23} \\
         \hline
         llm based chatbots for bpmn & 20 & process modelling, BPMN, BPM & 4 & \citep{bpmnentity},\citep{GPTMining},\citep{JsonAbs},\citep{VidgofBM23} \\
         \hline  
         prompt engineering and BPM & 13 300 & LLM, BPMN, BPM,  process modelling & 7& \citep{chit},\citep{Jana},\citep{PromptEng},\citep{VidgofBM23},\citep{GPTMining},\citep{LPM},\citep{DBLP:conf/bpm/KlievtsovaBKMR23} \\
         \hline      
\end{longtable}

\noindent\textbf{How are responses generated?} After receiving user input, the chatbot processes it into a machine-readable form and based on that input generates a natural language output utilising different types of response generation methods~\citep{response}. Chatbot systems can be divided into six categories, based on the type of response generator~\citep{types}.
   (1) template-based: response is selected from the list of predefined pairs of query patterns;
   (2) corpus-based: converts user query to a structured query language (SQL) query and passes it to utilised techniques of professional knowledge management (i.e., database, ontology);
   (3) intent-based: task-oriented system, which based on user query tries to recognise user intent with the help of advanced NLU techniques;
   (4) RNN-based: RNN-based (Recurrent Neural Network) chatbot generates response query directly from the user query with the help of the model, trained on dialogue dataset;
   (5) RL-based: RL-based (Reinforcement Learning) chatbots use rewarding and punishing functions to achieve the desired behaviour;
  (6) hybrid-based: a combination of approaches listed above to achieve better performance or to overcome limitations, faced by using one approach only.

\noindent\textbf{How can response generation be implemented?} All of the above types of response generator utilise some type of knowledge graph to formalise the configuration~\citep{graph,avila2020conquest} and the intended output format of the conversation~\citep{patsoulis2021integration,graph2}. The knowledge graph is either accessed by simple querying languages such as AIML or SPARQL, or it is encoded as part of a neural network through training. So responses are either queried explicitly or generated implicitly as part of
a neural network.
Both approaches have different strengths and weaknesses. For conversation-related applications such as entertainment, neural networks work well, but for other applications with special output, other approaches are still valid solutions. Low-code solutions to control explicit responses \citep{low-code} as well as BPMN-based solutions to encode potential progressions of a conversation~\citep{model} have been proposed. One example of such a system is Process-Aware Conversational Agent (PACA) \citep{pacas}.
Automatically learning from user interactions can be achieved not only for neural networks (e.g., reinforcement learning) but also by encoding interactions automatically into rules, such as in \citep{learnmodel,dpm}.

\noindent\textbf{Can chatbots deal with business process models?}\label{3tasks}
According to the survey of chatbot integration~\citep{survey}, 2 out of 347 chatbot systems support the business process interface pattern,
i.e., convert BPMN process models into dialog models/chatbots. Based on the business process model Quark generates an IBM Watson model for human-chatbot interactions required for the business process execution~\citep{quark}. ~\citep{modchat} introduces a methodology that transforms a BPMN model into a chatbot, which provides guidance and answers questions for individuals involved in this business process. 
Currently, there are no chatbots that are able to generate BPMN models themselves. However, interest in the generation of models from various types of document sources has recently increased~\citep{bpmngen,bpmngen3,bpmngen4}. Referring to~\citep{9946460} as an input for business process model generation use case diagrams, business rules, standard operating procedures, and plain unstructured text are considered.
Based on the approaches mentioned above, the following three steps for creating BPMN can be summarised~\citep{bpmnentity,8010715}: \phantomsection
\addcontentsline{toc}{section}{Some place in the document}
\label{some}
(1) Sentence Level Analysis: extraction of basic BPMN artefacts such as tasks, events, and actors;
(2) Text Level Analysis: exploration of relationships between basic items, e.g., gateways.
(3) Process Model Generation: create a syntactically correct model, that captures the semantics of the input.~\citep{atdp} proposes a machine-readable intermediate format generated out of natural language (either through automatic or manual annotation). The result is then easy to interpret by computers.

\noindent\textbf{Can AI chatbots empowered by LLMs deal with business process models?}
The recent development of LLMs and their integration into \textsl{conversational user interfaces} enable the development of a new type of LLM-based chatbots. As LLM capabilities expand from year to year, there is a growing interest in the potential benefits that can be derived for the entire BPM domain, e.g.,~\citep{LPM,PromptEng,ProcessGPT,VidgofBM23,chit}. 
In a more narrow context, LLM-based chatbots raised interest in process mining and process modelling, i.e., generation of new process models with event logs or unstructured text as input respectively. In~\citep{GPTMining}, for example, LLMs are applied for achieving an information abstraction of standard process mining artefacts such as traditional event logs, directly-follows graphs, or Petri nets.   
~\citep{bpmnentity} propose the extraction of process elements and relations using prompts with different levels of pre-knowledge. In ~\citep{Jana}, the generation of an entire model using a specific level of abstraction is presented. ~\citep{JsonAbs} utilises the JSON format to improve LLMs' capabilities to generate BPMN models, Entity-Relationship (ER), and UML Class Diagrams. ~\citep{bpmworkshop} uses a fine-tuned LLM for generating a recommendation for the next process element, i.e., its type and label in an unfinished BPMN process model. 

\noindent\textbf{How can we evaluate chatbots with respect to process modelling?}
\nk{It is challenging to establish benchmarks and metrics that provide a reasonable assessment of process extraction from unstructured text. Long time, there were no golden standard datasets that can be used to evaluate and compare the efficiency of process extraction from text~\citep{bpmngen2}. However, recently, a set of $47$ text-model pairs from industry and textbooks introduced in \citep{bpmngen} and annotated in~\citep{PETdata}, is considered as a benchmark. Additionally, $53$ model-text pairs are used in~\citep{modtext} to evaluate performance of a novel model-to-text transformation method. 
Data augmentation techniques can be used to increase the size of existing datasets with the help of the modified copies of already existing dataset items for training and fine-tuning purposes~\citep{augsur,methodology}.}  
Another important tool is paraphrasing~\citep{kauchak2006paraphrasing}, which is about generating similar texts from a source. Such texts are generally recognised as lexically and syntactically different while remaining semantically equal.

\srm{Section \ref{sec:sota} provides the results of a literature review and confirms and refines the taxonomy of application scenarios for ConverMod, referred to as Artefact 2 in Fig. \ref{fig:convermodprocessmethodabstract}. ConverMod is put to the test for these application scenarios in the subsequent sections, starting with the creation of a test set and a set of KPIs in Sect. \ref{sec:per}.} 

\section{\nk{Assessing Efficiency of Current Generation LLMs for Conversational Process modeling}}
\label{sec:per}
In order to assess the performance of ConverMod in a quantitative way, it is necessary to come up with a dataset, an evaluation method, and a set of KPIs. Extending the three steps, which are required to create a BPMN model (see Sect.~\hyperref[some]{3}), a fully integrated ConverMod toolchain is supposed to contain the following steps:
\setlist{nolistsep}
\begin{enumerate}[noitemsep]
 \item[(a)] extraction of tasks from textual descriptions;
 \item[(b)] extraction of logic such as decisions or parallel branching from textual descriptions (i.e., control flow); 
 \item[(c)] creation and the layout of a BPMN model; 
 \item[(d)] the application of modifications for refinement of BPMN models.
\end{enumerate}

As a fully integrated ConverMod tool does not exist yet. In this paper we focus on how well current LLMs perform for extracting tasks and control flow from textual description (see (a) and (b) above) and whether it is possible to obtain a reasonable model representation.

Each of the steps will be evaluated individually and only the models and prompts that achieved the best results will be considered for the next step.

\subsection{Test Set Generation}
\label{sec:texset}

\begin{figure}[ht]
    \centering
    \includegraphics[width=\textwidth]{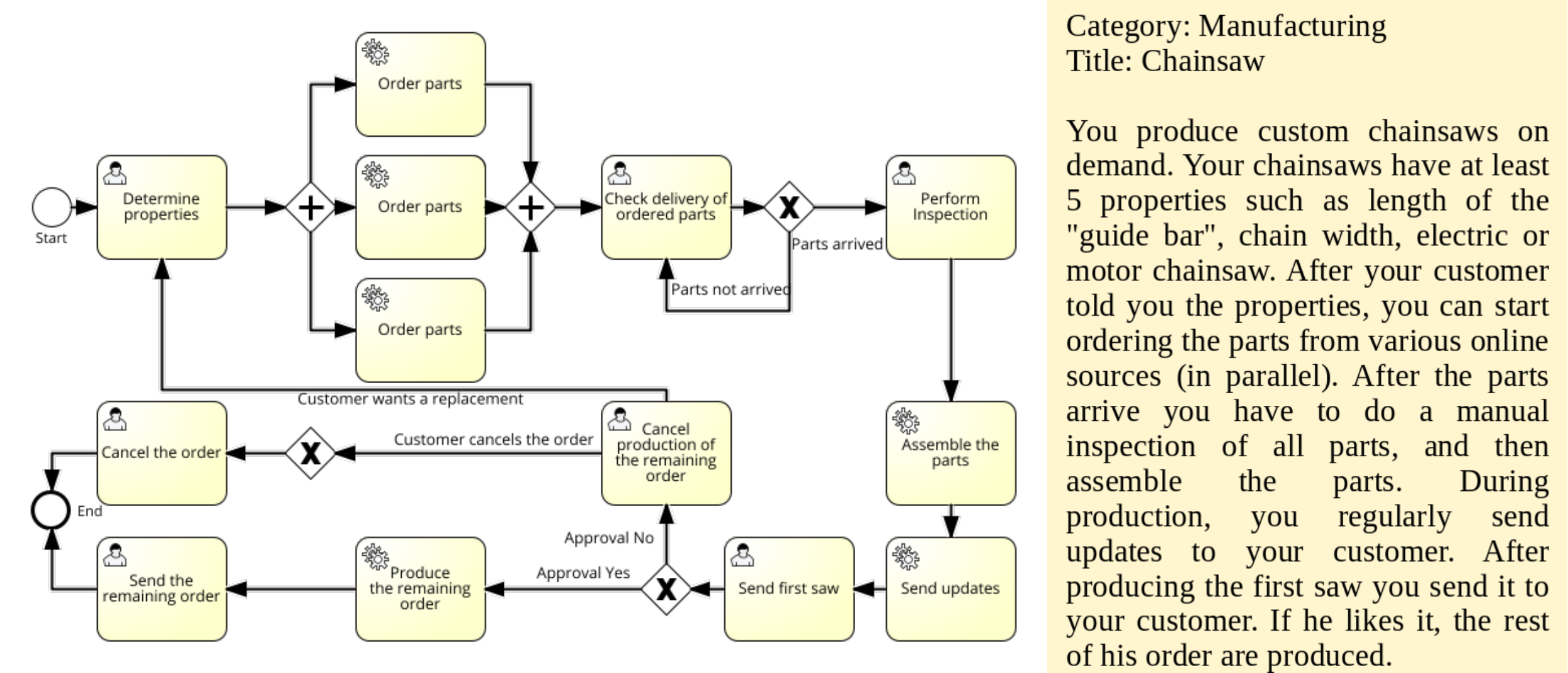}
    \caption{Textual Description And BPMN Model From the Evaluation Dataset}
    \label{fig:data_set}
\end{figure}

The test set\footnote{See~\citep{mangler_textual_2023}} utilised in this paper contains $21$
textual process descriptions from $6$ topics or domains ($\mapsto$ Artefact 3 in Fig. \ref{fig:convermodprocessmethodabstract}). For each
process description between $8$ and $11$ BPMN process models have been created by
modelling novices. These models represent different possible ways of interpreting the
textual process description. Each model has at least one start and end event, $3$
exclusive gateways, one parallel gateway, and an average of $14$ tasks. Some
models also contain sub-processes, pools, and lanes.
Each model was evaluated by a modelling expert using a quality value from $0$ to
$5$, to reflect, on how well the textual description has been transformed into a
BPMN model, i.e., all tasks and decisions from the textual description are in
the BPMN, tasks which can run in parallel have been correctly identified, and the
BPMN model is well-formed. 
An example of a textual description and an associated interpretation, i.e., the BPMN model, can be seen in Fig.~\ref{fig:data_set}.

\subsection{\srm{Quantitative Evaluation: Assessing Process Model Completeness}}
\label{sec:eval}
\srm{In this section, we perform task and control flow extraction with ConverMod on the test set described in Sect.~\ref{sec:texset} and propose a set of KPIs ($\mapsto$ Artefact 4) to evaluate the results.}

\subsubsection{Task Extraction}
\label{sec:taskextraction}
Task extraction is the starting point of the ConverMod toolchain, as the task is an atomic element of the process flow, which represents a unit of work that should be performed~\citep{atomic}.

For assessing the efficiency of extracting tasks from textual process descriptions specific quality metrics (i.e., KPIs) are required. Several works have addressed \jvb{quantifying} the quality of the model~\citep{quality1,quality2}. In the majority of cases, during the evaluation of a model's quality, specific parameters are assessed in accordance with predefined guidelines to ensure clarity and comprehension.

However, there is a gap in the assessment of how close a particular model is to another one or its representation in another format. The significance of existence and the advantages offered to the organization, along with the accompanying challenges are listed in~\citep{fragmentation}. Only a few works have tackled this problem. For instance, a similarity measure between two models is proposed in~\cite{bpmngen} and to measure the  distance between sentence and activity semantic, \citet{inconsist} use a similarity score. 

Neither of the proposed methods is applicable to our particular case, as the goal is to estimate the level of completeness captured by extracted tasks  concerning the original process description. Even though the model created by a stakeholder could have some deviations from the process description created by another stakeholder due to terminology, structure and abstraction level ~\citep{fragmentation}, we can suppose that the proposed model contains all information related to the process description. Nonetheless, in the scenario, where a process model (i.e., a list of \jvb{tasks}) is created by a LLM, such an assumption \jvb{cannot be justified}. 

In this section, to evaluate how effective task extraction from textual process description is the following LLMs, namely GPT models are used: text-davinci-001 (GPT1), text-davinci-002 (GPT2), text-davinci-003 (GPT3), gpt-3.5-turbo (GPT3.5) and gpt-4 (GPT4) from openai.org\footnote{last access: 2023-10-25}. We will propose to utilize the following KPIs and discuss their impact on ConverMod approaches:
\textbf{KPI1} - Text Similarity;
\textbf{KPI2} - Set Similarity;
\textbf{KPI3} - Set Overlap;
\textbf{KPI4} - Restricted Text Similarity;
\textbf{KPI5} - Restricted Set Similarity;
\textbf{KPI6} - Restricted Set Overlap;
\textbf{KPI7} - Average Augmented Task Extraction Prevalence and Similarity (GPT3 and GPT4 only).
All results, including non-averaged data, are also available in git repository\footnote{\label{arepo} See~\url{https://github.com/WSaccoun/convermod}}.

KPIs 1--3 are used to assess task extraction from the original process descriptions. This is realized by passing the following prompt ``Considering following $<process\_description>$ return the list of main tasks in it'' to the LLMs. For assessing KPIs 4--6, the original prompt is changed to the ``Considering following $<process\_description>$ return the list of main tasks (each 3-5 words) in it'' to improve the granularity of extracted tasks and to refine the quality of obtained tasks' labeling. KPI7 is used to evaluate how stable task extraction, performed by LLMs, is by extending the set of original process descriptions by utilizing different paraphrasing algorithms. 

\begin{figure}[ht]
    \centering
    \includegraphics[width=1\textwidth]{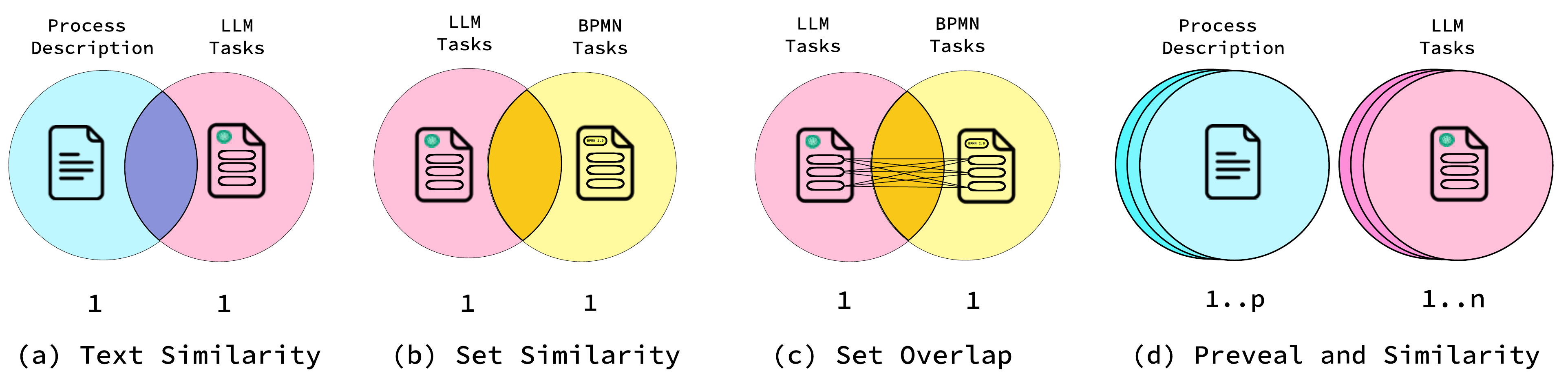}
    \caption{Key Performance Indicators (KPIs)}
    \label{fig:kpi}
\end{figure}

Task extraction from associated models is realised by parsing XML documents and extracting relevant BPMN activities, keeping their sequence in the process flow. 

As the basis for each similarity measurement, we utilize contextual (BERT) and non-contextual (TD-IDF) vectorisers with a cosine similarity metric~\citep{chandrasekaran_evolution_2021}. The contextual and non-contextual approaches will be denoted as C and NC.

For \textbf{KPI1}, each LLM (GPT1, GPT2, GPT3, GPT3.5, GPT4) is instructed to extract
the tasks from the original process descriptions. The answer is then compared to the original text to assess the completeness of the extraction (see Fig.~\ref{fig:kpi}). The results are depicted in
Table~\ref{tab:textsim}. For this KPI, GPT3.5 is the most successful LLM.

\begin{table}[!htp]
    \centering
    \caption{Text Similarity (\textbf{KPI1}): Comparison of tasks extracted by LLM and original text using contextual (C) and non-contextual (NC) vectorisers}
  
        \begin{tabular}{c|c|c|c|c|c}
         \textbf{Method} & \textbf{GPT1} & \textbf{GPT2} & \textbf{GPT3}  & \textbf{GPT3.5} & \textbf{GPT4}  \\\hline  %
         NC & 0.46	& 0.65 &	0.60    &  0.63 & 0.62  \\\hline   %
         C  & 0.76	& 0.80	  & 0.78	& 0.84   & 0.82\\         %
    \end{tabular}

    \label{tab:textsim}
\end{table}

Table~\ref{tab:setsim} shows the results for \textbf{KPI2}. The five LLMs
are instructed to extract tasks from each textual description. This set of
tasks is then compared to the set of tasks that has been extracted from each BPMN model mentioned above (see Sect.~\ref{sec:texset}). As for every textual description multiple BPMN models
exist, the results are averaged per textual description. The averages are then
again averaged for all textual descriptions(See Fig.~\ref{fig:kpi}). Both GPT3 and GPT4 are the \jvb{most} successful for this KPI with 74\%
extraction rate.

\begin{table}[htb!]
    \centering
    \caption{Set Similarity (\textbf{KPI2}): Comparison of tasks extracted by LLM with tasks extracted from BPMN Models. For each text a set of n tasks is extracted. Each text has 8--11 associated models from which again a set of m tasks can be extracted. Each set n is compared with all sets m, yielding a set of similarities which is averaged for similarity methods contextual (C) and non-contextual (NC)}
      \begin{tabular}{c|c|c|c|c}
          \multirow{2}{4em}{\textbf{LLM}} &  \multirow{2}{4em}{\textbf{C}} & \multirow{2}{4em}{\textbf{NC}} & \multirow{2}{10em}{\textbf{avg. \# of tasks extracted from texts }}  & \multirow{2}{10em}{\textbf{avg. \# of tasks extracted from models} } \\ &&&&
            \\\hline
          GPT1 & 	 0.72 & 0.32 &	7.6	& 12  \\\hline
          GPT2 & 	 0.71 & 0.32 &	6.7	& 12  \\\hline
          GPT3 & 	 0.74 & 0.35 &	7.7	& 12  \\\hline
          GPT3.5 & 0.73 & 0.36 &	8.5	& 12  \\\hline
          GPT4 & 	 0.74 & 0.35 &	9.7	& 12  \\
      \end{tabular}
     \label{tab:setsim}
\end{table}

For \textbf{KPI3}, the goal is to quantify the overlap between extracted tasks from the original process description with LLM and \nk{tasks extracted from associated to this description models}: (1) how similar are individual tasks, and (2) how many tasks exist only in one of the two extractions (See Fig.~\ref{fig:kpi}). The results are shown in Table~\ref{tab:setoverlap} and demonstrate that between $6$ and $7$ tasks extracted from the models are also found in the text, while about 6 tasks could not be found in the extracted text. When looking at it from the point of view of the tasks extracted from the text, the ratio becomes 4:3. So almost 50\% of the tasks are not similar between the model and original text (see discussion for details).

\begin{table}[htb!]
    \centering
    \caption{Set Overlap (\textbf{KPI3}): Each task extracted from the text is compared (for each associated model) with task extracted from the model. If the similarity is bigger than a threshold, a task is deemed common, else it is deemed to only occur in either the model or the text.}
        \begin{tabular}{c|c| c|c|c|c}
         \textbf{LLM} & \textbf{similarity} & \textbf{common model}  & \textbf{common chat} & \textbf{only in model}  & \textbf{only in chat}\\\hline
         GPT1 & 	C  &  6.6  & 4.5 & 5.2 & 3.2  \\\hline
         GPT1 & 	NC  & 5.9  & 4 & 5.9 & 3.6  \\\hline
         GPT2 & 	C  &  6.2  & 4.1 & 5.6 & 2.6  \\\hline
         GPT2 & 	NC &  5.6  & 3.6 & 6.2 & 3   \\\hline
         GPT3 & 	C  &  6.7  & 4.6 & 5.1 & 3  \\\hline
         GPT3 & 	NC &  6.7 & 4.6 & 5.1 & 3   \\\hline
         GPT3.5 & 	C &   7 & 4.7 & 4.9 & 3.8   \\\hline
         GPT3.5 & 	NC &  6.5 & 4.4 & 5.4 & 4.1   \\\hline
         GPT4 & 	C &   7.5 & 5.2 & 4.3 & 4.4   \\\hline
         GPT4 & 	NC &  6.8 & 4.9 & 5.0 & 4.8   \\
    \end{tabular}
    \label{tab:setoverlap}
\end{table}

\textbf{KPI4} focuses on restricting the number of words per extracted task,
to coax the bot into extracting more tasks, as generally, the number of
extracted tasks from the text is lower than the number of tasks contained in
the models (see discussion for more details). Table~\ref{tab:resttextsim} shows
that this decreases the similarity when comparing text (due to stronger
paraphrasing), but \textbf{KPI5} (cf. Table \ref{tab:restsetsim}) and  \textbf{KPI6} (cf. Table \ref{tab:restsetoverlap}) show an increase in the
number of tasks by one while not decreasing similarity when compared to the tasks
from the model.

\begin{table}[htb!]
    \centering
    \caption{Restricted Text Similarity (\textbf{KPI4}): Task names are allowed to only have 3-5 words, cmp. Table~\ref{tab:textsim}.}
         \begin{tabular}{c|c|c|c|c|c}
         \textbf{Method} & \textbf{GPT1} & \textbf{GPT2} & \textbf{GPT3}  & \textbf{GPT3.5} & \textbf{GPT4} \\\hline   %
         non-contextual & 0.24	&	0.47	&	0.38 & 0.27  & 0.29\\\hline %
         contextual & 0.70	&	0.77	&	0.73 & 0.73  & 0.76   \\    %
    \end{tabular}
    \label{tab:resttextsim}
\end{table}

\begin{table}[htb!]
    \centering
    \caption{Restricted Set Similarity (\textbf{KPI5}): Task names are allowed to only have 3-5 words, cmp. Table~\ref{tab:setsim}.}
      \begin{tabular}{c|c|c|c|c}
          \multirow{2}{4em}{\textbf{LLM}} &  \multirow{2}{4em}{\textbf{C}} & \multirow{2}{4em}{\textbf{NC}} & \multirow{2}{10em}{\textbf{avg. \# of tasks extracted from texts }}  & \multirow{2}{10em}{\textbf{avg. \# of tasks extracted from models} } \\ &&&&
            \\\hline
          GPT1 & 	 0.73 & 0.32 &	7.6	& 12  \\\hline
          GPT2 & 	 0.74 & 0.33 &	7.7	& 12  \\\hline
          GPT3 & 	 0.73 & 0.32 &	8.3 & 12  \\\hline
          GPT3.5 & 0.75 & 0.30 &	8.5	& 12  \\\hline
          GPT4 & 	 0.77 & 0.33 &	11.6& 12  \\
      \end{tabular}
    \label{tab:restsetsim}
\end{table}

\begin{table}[htb!]
    \centering
    \caption{Restricted Set Overlap (\textbf{KPI6}): Task names are allowed to only have 3-5 words, cmp. Table~\ref{tab:setoverlap}.}
          \begin{tabular}{c| c| c|c|c|c}
         \textbf{LLM} & \textbf{similarity} & \textbf{common model}  & \textbf{common chat} & \textbf{only in model}  & \textbf{only in chat}\\\hline
         GPT1 & 	NC  & 6  & 4 & 5.7 & 3.5  \\\hline
         GPT2 & 	NC &  6.4  & 4.2 & 5.4 & 3.5   \\\hline
         GPT3 & 	NC &  7 & 4.7 & 4.8 & 3.5   \\\hline
         GPT3.5 & 	NC &  6.8 & 4.6 & 5 & 3.8 \\\hline
         GPT4 & 	NC &  7.8 & 5.6 & 4 & 6
    \end{tabular}
    \label{tab:restsetoverlap}
\end{table}

Finally, for \textbf{KPI7}, we assess the effects of paraphrasing on prevalence and similarity, i.e., how stable LLMs are for task extraction with similar input. We use nine different algorithms for paraphrasing text~\citep{augsur} (i.e., rewriting sentences using synonyms), which is, for example, useful to clean up textual descriptions from humans. The results are displayed in Table~\ref{tab:augmented}, and show that especially contextual similarity does not decrease substantially, while the number of extracted tasks even improves in comparison to the original text.

\begin{table}[htb!]
    \centering
    \caption{Average Augmented Task Extraction Prevalence and Similarity for GPT3 and GPT4 (\textbf{KPI7}): for nine different paraphrasing methods, the average number of tasks, and similarity measures are calculated. The second row holds the value of the original text from Table~\ref{tab:resttextsim}}
        \begin{tabular}{c|c| c| c|c|c|c|c|c|c|c}
         & \textbf{Original} & \textbf{SR} & \textbf{DL}  & \textbf{SW} & \textbf{IN}  & \textbf{NLPaug} & \textbf{TDE} & \textbf{TRU}  & \textbf{TES} & \textbf{EMB}\\\hline
         \multicolumn{11}{c}{GPT3}\\\hline
         avg. \# tasks & 8.25	&	8.10	& 8.43	& 7.48	& 8.19	& 8.10 &	7.57	& 7.86	& 8.62 & 8.29 \\\hline
         C similarity  & 0.73	&	0.69&	0.69&	0.68	&0.70&	0.70&	0.70&	0.67&	0.70&	0.70 \\\hline
         NC similarity & 0.38	&	0.20	& 0.22 &	0.25&	0.21&	0.21&	0.21&	0.19&	0.21&	0.22 \\\hline
         \multicolumn{11}{c}{GPT4}\\\hline
         avg. \# tasks & 11.7 & 12.19	&11.29	&11.95	&11.10	&11.67	&12.14	&11.86	&11.62	&11.71 \\\hline
         C similarity &0.76  & 0.73	&0.74	&0.74	&0.74	&0.73	&0.74	&0.70	&0.74	&0.73\\\hline
         NC similarity & 0.26 &0.26	&0.29	&0.28	&0.30	&0.26	&0.26	&0.22	&0.23	&0.26
    \end{tabular}
    \label{tab:augmented}
\end{table}

\textbf{Summary.} In general, tables \ref{tab:textsim} to \ref{tab:augmented} demonstrate no significant difference in results between all GPT models. However, \textbf{KPI5} and \textbf{KPI6} (see Tables~\ref{tab:restsetsim},~\ref{tab:restsetoverlap}) clearly show that GPT4 currently supports task extraction the best, beating other models. By restricting the number of words in task labels, GPT4 was able not only to extract, on average, almost the same number of tasks as humans did but also to reach the highest level of Restricted Set Overlap (\textbf{KPI6}) between the provided process description and its associated models. The reason could be that GPT4 is the largest model with over 1.76 trillion parameters.

GPT3 is considered the second-best model. Even though the number of extracted tasks using GPT3 is slightly smaller than when utilising GPT3.5, the \textbf{KPI6} of GPT3 is better than that of other models. The potential explanation for this is the ability of GPT3 to handle more complex instructions. The GPT3.5 model, on the other hand, is optimised for chat and may not be as effective for more advanced language tasks~\citep{openai}.

GPTs can only extract tasks included in the original process description. On the other hand, modellers tend to add new information to models (see Sect.\ref{sec:disc} for more details).

\subsubsection{Control Flow Extraction}
\label{sec:controlflow}
Due to results from Sect.~\ref{sec:taskextraction}, only GPT4 and GPT3 and restricted labels \jvb{are} considered for future assessment.

In order to evaluate the ability of LLMs to extract the control flow from the provided text description, we \jvb{ask} them to create BPMN models and then compare the received models to the existing standard. It is necessary to consider both, quantitative and qualitative factors of the produced models (i.e., process description content encompassing, similarity to a standard, model correctness) to determine the accuracy of LLMs to generate a process model.

Generally, the whole assessment procedure comprises of three steps: prompt engineering, selection of textual representations, and estimation of produced graphs.  

\textbf{Prompt Engineering}. The following prompts are designed to pass process descriptions and to provide additional context, e.g., information about the BPMN standard, custom rules of textual notation for \nk{selected textual-based graph representations}, or a list of tasks that were extracted from it to LLMs to guide them through the process of model generation:
 \begin{enumerate}
     \item[(S)] process description only;
     \item[(A)] process description and a list of tasks extracted from it;
     \item[(R)] process description and a set of custom rules for selected textual graph representation;
     \item[(B)] process description and description of utilised BPMN elements;
     \item[(RA)] process description, a list of extracted tasks and a set of custom rules for selected textual graph representation;
     \item[(RAB)] process description, a list of extracted tasks, a set of custom rules for selected textual graph representation and a description of utilised BPMN elements.
 \end{enumerate}
 The prompts, a short description of custom textual notation for selected text-based graph representation and a simplified description of utilised BPMN elements according to BPMN 2.0 standard can be found here: (see \footref{arepo}).

\textbf{Selection of Graphical Representations}. Due to the context length limitation of LLMs, it is not possible to utilise standard XML serialisation of BPMN models. For instance, a regular language model input/output limit is between 1.000-8.000 tokens. At the same time, a simple BPMN model with 7 tasks in XML representation consists of 10.000 to 15.000 tokens, depending on its complexity. Therefore, it is required to use an abstract graphical representation of the BPMN model. Multiple comprehensible graphical representations (GR) of various types of event logs or process models (e.g., directly-follows graph (DFG) or Petri net) are suggested in~\citep{Abstractions} and pre-specified intermediary textual notation (TN) of BPMN are proposed in~\citep{Jana}. However, ordinary TNs not only requires further conversion into diagrams (i.e., GNs), but also don't allow us to see and analyse the output of a language model immediately in a user-friendly/understandable manner. 

Nowadays, multiple TNs like DOT\footnote{https://graphviz.org/doc/info/lang.html}, Mermaid.js\footnote{https://mermaid.js.org/}, Penrose\footnote{https://penrose.cs.cmu.edu/}, D2\footnote{https://d2lang.com/tour/intro} and PlantUML\footnote{https://plantuml.com/} exist\citep{dialang}. As some of these languages are more diagram type specific and not so generic, similar to \citep{bpmworkshop}, Mermaid.js (MER) was selected to create BPMN diagrams by using their textual definition. MER is intuitive and easy to integrate into applications. The tool that converts the MER TN into BPMN GN is called BPMN2Constraints \footnote{https://github.com/signavio/bpmn2constraints}. In addition, we use the TN Graphviz DOT (GV) as an alternative way to create a graphical BPMN process model, as it is widely used, stable, well-documented, and can be integrated into various programming languages and tools. 

\begin{figure}[ht]
    \centering
    \includegraphics[width=0.9\textwidth]{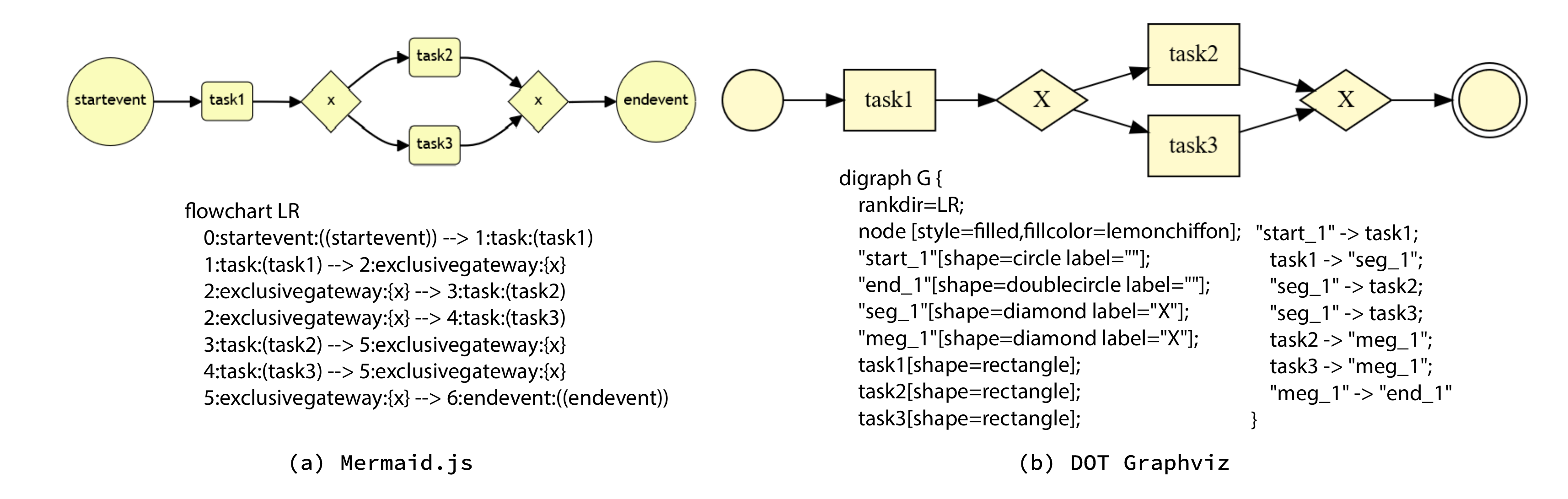}
    \caption{TN \& GN of Text Fragment given in TF\ref{q:simple}}
    \label{fig:mervsgv}
\end{figure}

\nk{The TN contains the following elements: At first, we define the orientation of the graph and custom structure for all nodes and edges in it. Then, specific features are assigned for every node to represent particular BPMN element.}  For example the simple process description 
\begin{tfquote}
"After task1, either task2 or task3 are conducted."     
\label{q:simple}
\end{tfquote}
can be converted by a LLM into the TN and finally into the GN shown in Figure~\ref{fig:mervsgv} (for more details see~\footref{arepo}).

\textbf{Graph Estimation}. As GPT models can solely operate with information existing in the original text, for this section, the PET dataset\footnote{https://huggingface.co/datasets/patriziobellan/PET}~\citep{PETdata} is used as a standard (i.e., the ground truth or correct process model). Based on the existing human annotation of activities, gateways, and control flow provided in the PET dataset, BPMN process models were created manually. 
Seven examples are taken over from~\cite{bpmngen2}, as they are of different lengths and complexity. \jvb{The variety in length and complexity is crucial}, because the prompt's length and content supplied to the LLM directly affect the quality of the generated response~\cite{respgen}. 

For an evaluation process, two basic categories of BPMN elements are considered: (a) flow objects (start and end events, tasks, exclusive and parallel gateways) and (b) connecting objects (sequence flows). The concept of merging gateways is addressed as follows: if the LLM does not generate a merging gateway, we do not consider it an error. A merging gateway concept is not a strict principle of the BPMN standard and is often ignored by human modellers. If two tasks are merged into one, we consider it as a correct LLM response and split it manually into distinct tasks (e.g., ``check and repair hardware'' is equal to a sequence of two tasks ``check hardware'' and ``repair hardware''). If the textual representation was generated with some syntactical errors, this was manually adjusted, but still taken into consideration.  

With $6$ existing prompts and $7$ selected examples, both GPT3 and GPT4 are instructed to create MER and GV graphs. XML representations of standard BPMN models and LLM-generated graphs are transformed into a list of 3-tuples, where every tuple is an ordered sequence of BPMN objects (target flow object, connecting object, source flow object). One tuple represents an edge of a model/graph. 

\begin{figure}[ht]
    \centering
    \includegraphics[width=0.7\textwidth]{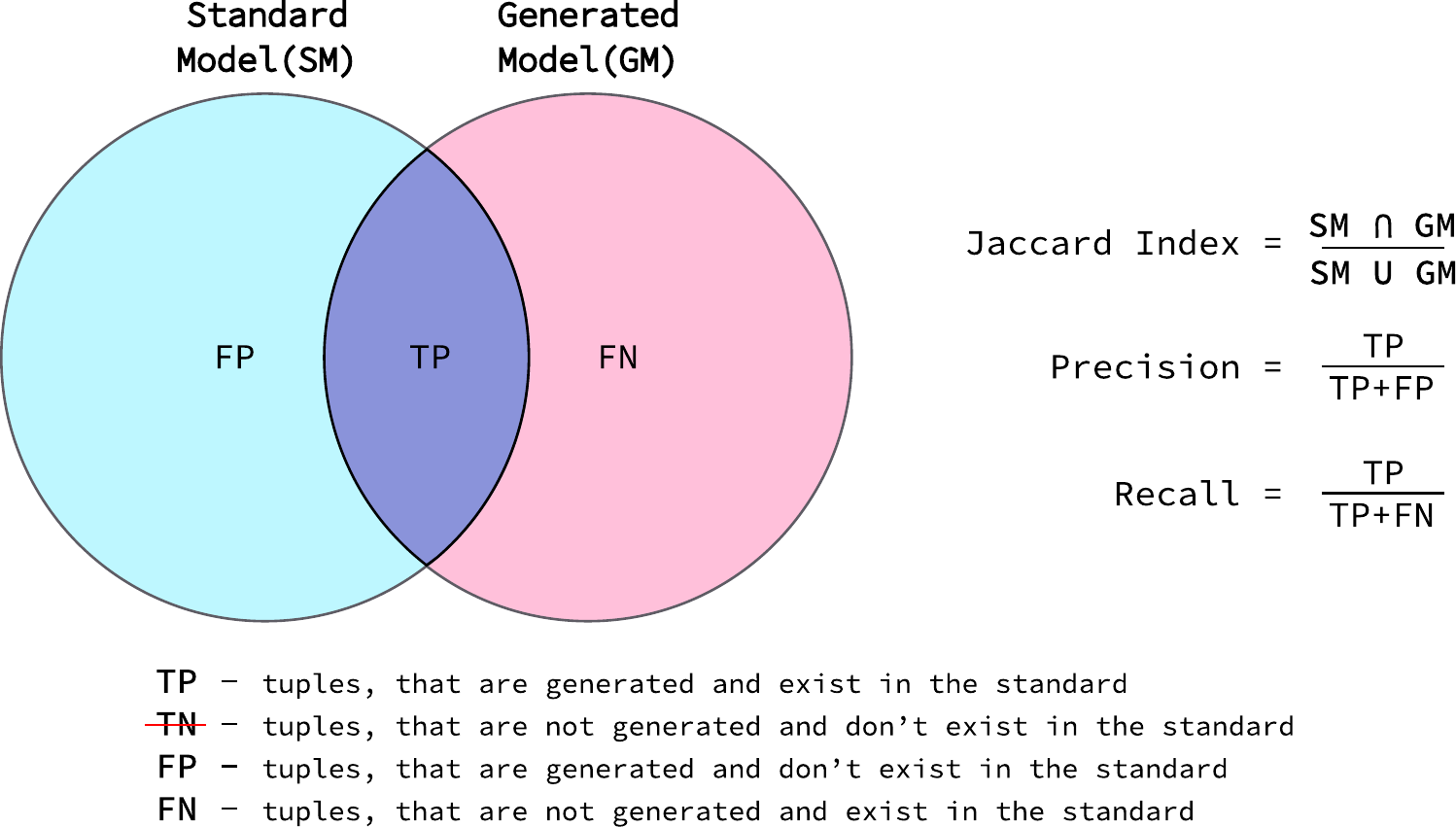}
    \caption{Quantitative Evaluation Metrics for Generated Models: Jaccard Index, Precision, Recall}
    \label{fig:jipr}
\end{figure}

\textbf{Model Completeness}. 
The following metrics are selected to \jvb{quantitatively assess the} generated models, i.e., the Jaccard coefficient, precision and recall. 
We use the Jaccard index to measure the similarity between the \jvb{model} generated by the LLM model and the original \jvb{model} (see Fig. \ref{fig:jipr}). It allows us to determine the similarity between two sets without considering the sequence order of the items in a sample set. 

We \jvb{compute} precision and recall metrics to \jvb{assess the overall quality of the generated models}. In the NLP domain, precision and recall are used for classification problems or information retrieval evaluation ~\citep{metrics}. This paper does not deal with a standard classification problem where the goal is to select values that closely match the target. We are dealing with a specific case where the main goal is to generate a new set of values resembling the target. 

The following parameters are used: (tp) are generated tuples that exist in the standard models as well; (fp) are tuples that were generated but do not exist in the standard models; and (fn) are tuples that were not generated but still exist in the standard model. \jvb{Consequently}, (tn) values are always $0$, i.e., do not exist as these tuples are not generated and are not mentioned in the standard (See Fig.~\ref{fig:jipr}).

Precision is then a measure of the proportion of correctly generated tuples out of all generated tuples, and recall is the proportion of correctly generated tuples out of all target tuples~\citep{metrics}.

\begin{table}[!ht]
    \centering
    \caption{Average Jaccard Index for PET standard models and textual notations (TNs) generated with GPT3 and GPT4; abbreviations see above}
    \begin{tabular}{l|l|l|l|l|l|l|l||l|l}
        \textbf{LLM} & \textbf{TN} & \textbf{S} & \textbf{A} & \textbf{R} & \textbf{RA} & \textbf{RAB} & \textbf{B} & \textbf{F. Human} & \textbf{F. System} \\ \hline
        \multirow{2}{*}{GPT3} & GV & 0.03 & 0.11 & 0.44 & 0.47 & 0.15 & 0.03 & 0.47 & 0.58 \\ 
                              & MER & 0.18 & 0.22 & 0.52 & 0.47 & 0.39 & 0.19 & 0.47 & 0.58 \\ \hline
        \multirow{2}{*}{GPT4} & GV & 0.13 & 0.17 & 0.50 & 0.56 & 0.46 & 0.40 & 0.47 & 0.58 \\ 
                              & MER & 0.14 & 0.22 & 0.54 & 0.57 & 0.59 & 0.27 & 0.47 & 0.58 \\ 
    \end{tabular}
    \label{tab:jacc}
\end{table}

In Table ~\ref{tab:jacc} the results for the Jaccard Index (JI) are presented. For a benchmark, we consider the models created by a human modeller and an automated system from~\citep{friedrich}. As we can see, by using prompts with only the process description or with the process description and additional information that does not provide sufficient instructions about the textual representation format (i.e., (S), (A) or (B)), the JI value does not reach more than 27\% in the best case (20\% on average). 

For the rest of the cases, using GPT3, a maximum of $52$\% similarity was achieved. However, there is a $13$\% decrease between models built with (R) and (RAB) prompts.  This might suggest that the more content is supplied, the worse the LLM's understanding of the instructions is.

By generating models with GPT4, we attain JI values of $59, 57$\% and $54$\% with (RAB), (RA) and (R) prompts, respectively. These results are similar to those reached by the automated system and $10$\% higher than the result obtained by a human modeller from the benchmark.    

\begin{table}[!ht]
    \centering
    \caption{Average Precision for PET standard models and textual notations (TNs) generated with GPT3 and GPT4; abbreviations see above}
    \begin{tabular}{l|l|l|l|l|l|l|l||l|l}
        \textbf{LLM} & \textbf{TN} & \textbf{S} & \textbf{A} & \textbf{R} & \textbf{RA} & \textbf{RAB} & \textbf{B} & \textbf{F. Human} & \textbf{F. System} \\ \hline
        \multirow{2}{*}{GPT3} & GV & 0.06 & 0.24 & 0.57 & 0.58 & 0.26 & 0.07 & 0.38 & 0.65 \\ 
                              & MER & 0.37 & 0.45 & 0.70 & 0.60 & 0.55 & 0.38 & 0.38 & 0.65 \\ \hline
        \multirow{2}{*}{GPT4} & GV &  0.28 & 0.36 & 0.60 & 0.66 & 0.53 & 0.48 & 0.38 & 0.65 \\ 
                              & MER & 0.32 & 0.43 & 0.65 & 0.67 & 0.70 & 0.38 & 0.38 & 0.65 \\ 
    \end{tabular}
    \label{tab:precision}
\end{table}

Similar to the JI results (see Table~\ref{tab:jacc}), Table~\ref{tab:precision} shows that when providing less informative instructions to LLMs models, models with lower precision are created. Only between $32$--$45$\% of all generated tuples are correct. Nonetheless, models designed by human modeller score $38$\% precision compared with the standard. 

With more content-substantial prompts, mostly $53$---$70$\% precision is reached. \jvb{Hence, generated model precision} is $5$\%-points better compared to the benchmark with the automated system and almost $2$ times better \jvb{compared} to the human-designed models. 

Table~\ref{tab:recall} presents recall values to identify how many target tuples were considered in the generated models.
In most cases, using (R),(RA) and (RAB) prompts, a recall of $51$--$71$\% is achieved. That is similar to the benchmark, where $58$\% and $69$\% recall was obtained by a human modeler and an automated system, respectively. Executing less descriptive prompts, only between $5$\% and $30$\% of target tuples were covered by generated models.  

\begin{table}[!ht]
    \centering
    \caption{Average Recall for PET standard models and textual notations (TNs) generated with GPT3 and GPT4; abbreviations see above}
    \begin{tabular}{l|l|l|l|l|l|l|l||l|l}
        \textbf{LLM} & \textbf{TN} & \textbf{S} & \textbf{A} & \textbf{R} & \textbf{RA} & \textbf{RAB} & \textbf{B} & \textbf{F. Human} & \textbf{F. System} \\ \hline
        \multirow{2}{*}{GPT3} & GV & 0.04 & 0.14 & 0.53 & 0.58 & 0.26 & 0.05 & 0.58 & 0.69 \\ 
                              & MER & 0.23 & 0.28 & 0.57 & 0.57 & 0.51 & 0.25 & 0.58 & 0.69 \\ \hline
        \multirow{2}{*}{GPT4} & GV & 0.20 & 0.25 & 0.62 & 0.70 & 0.61 & 0.51 & 0.58 & 0.69 \\ 
                              & MER & 0.19 & 0.30 & 0.65 & 0.70 & 0.71 & 0.30 & 0.58 & 0.69 \\ 
    \end{tabular}
    \label{tab:recall}
\end{table}

\textbf{Summary.} In general, better results are achieved by applying prompts with more explanatory content, i.e., (R), (RA), and (RAB). Results of JI over $50$\% along with both recall and precision exceeding $70$\% can be considered satisfactory. However, there is room for further improvement.

Observing all three metrics, a consistent pattern is found: supplying GPT3 and GPT4 with less content leads to \jvb{worse} results. \jvb{At the same time}, supplying GPT3 with too much content leads to a poorer understanding of the provided instructions (decrease in performance by (RA) and (RAB) in comparison to (R) prompts). \jvb{Interestingly}, feeding GPT4 with more content leads to better results (i.e., (RA) and (RAB) achieve better scores as (R) prompts).  

Analysing Table~\ref{tab:jiprsummary}, we can see that utilising MER GPT3 was able to achieve $15$--$20$\%\jvb{-points} better scores \jvb{compared to }GV. GPT4 shows only a 5\%\jvb{-points} increase in performance in contrast to GPT3 with MER, but at the same time reaches $20$\%\jvb{-points} better results using GV. However, there is a slight difference ($3$\%\jvb{-points}) between both text-based graph representations applying GPT4.

\begin{table}[!ht]
    \centering
    \caption{Average of average: Jaccard Index, Recall, Precision for Models and Tasks; abbreviations see above}
    \begin{tabular}{l||l|l|l|l|l|l||l|l|l|l|l|l}
        Metric & \multicolumn{2}{c|}{JI} & \multicolumn{2}{c|}{Precision} & \multicolumn{2}{c||}{Recall}   & \multicolumn{2}{c|}{JI} & \multicolumn{2}{c|}{Precision} & \multicolumn{2}{c}{Recall}  \\ \hline
        TN & GV & MER & GV & MER & GV & MER  & GV & MER & GV & MER & GV & MER  \\ \hline
        ~ & \multicolumn{6}{c||}{Models} & \multicolumn{6}{c}{Tasks} \\ \hline
        GPT3 & 0.21 & 0.33 & 0.3 & 0.51 & 0.2 & 0.4 & 0.51 & 0.68 & 0.73 & 0.85 & 0.57 & 0.76 \\ \hline
        GPT4 & 0.37 & 0.39 & 0.48 & 0.53 & 0.48 & 0.47 & 0.71 & 0.69 & 0.78 & 0.84 & 0.85 & 0.78  \\ 
    \end{tabular}
    \label{tab:jiprsummary}
\end{table}

In addition, we considered the same metrics for the task extraction evaluation by taking out the set of tasks from the standard models and models created by LLMs. Despite relatively low JI values of models in general (between $21$ and $39$\%), executing different prompts, LLMs are able to achieve $51$--$71$\% JI when comparing task sets. Recall and precision for the task extraction reach up to $85$\% on average, and are mostly similar or even better than by the human modeller and an automated system from the benchmark (see Table~\ref{tab:jiprsummary} and~\footref{arepo}).

\nk{Along with newly introduced above metrics we take a look on KPIs (i.e., KPI 4-6) mentioned in Sect.~\ref{sec:taskextraction}. The winner is GPT4. Using MER TN, a GPT4 NC and C Text Similarity of $56$ and $85$\% is reached. Semantic similarity between standard models and tasks extracted from process description by LLMs is $78$\% on average with the most supreme result of $90$\% utilizing GPT4 and MER.}

\begin{table}[!ht]
    \centering
    \caption{Restricted Set Overlap (\textbf{KPI6}), cmp. Table~\ref{tab:restsetoverlap}.}
    \begin{tabular}{l|l|l|l|l|l|l|l|l}
        LLM & TN & Method & com. model & com. llm & only model & only llm & llm tasks & bpmn tasks \\ \hline
        GPT3 & GV & C & 4.76 & 3.95 & 1.81 & 1.24 & 5.19 & 6.57 \\ \hline
        ~ & MER   & C & 4.43 & 3.43 & 2.14 & 1.24 & 4.67 & 6.57 \\ \hline
        GPT4 & GV & C & 5.14 & 4.29 & 1.43 & 1.76 & 6.05 & 6.57 \\ \hline
        ~ & MER   & C & 6.33 & 4.95 & 0.24 & 1.24 & 6.19 & 6.57 \\ \hline
        AVG & ~ & ~ & 5.17 & 4.15 & 1.40 & 1.37 & 5.52 & 6.57 \\ \hline
        (\%) & ~ & ~ & 79 & 75 & 21 & 25 & ~ & ~ \\ 
    \end{tabular}
    \label{tab:kpi6-pet}
\end{table}

\nk{High precision and recall for the tasks could be explained by Set Overlap values. Up to $79$\% of all tasks from the standard model are found in models generated by LLMs and $75$\% out of all LLMs generated tasks exist in the standard model (see Table~\ref{tab:kpi6-pet}). For simple process descriptions, LLMs tend either to identify all activities and relations correctly or create additional elements, that are not considered as ones. For more complex process descriptions, LLMs use to ignore up to $30$\% of original activities on average (see~\footref{arepo}). This leads to low precision and recall values for model's relations (see Table~\ref{tab:jiprsummary}, as one skipped activity or gateway causes high level of misalignment between two models.}

\subsection{\srm{Qualitative Evaluation: Assessing Process Model Correctness}}
\label{sub:correctness}
\begin{figure}[ht]
    \centering
    \includegraphics[width=1.0\textwidth]{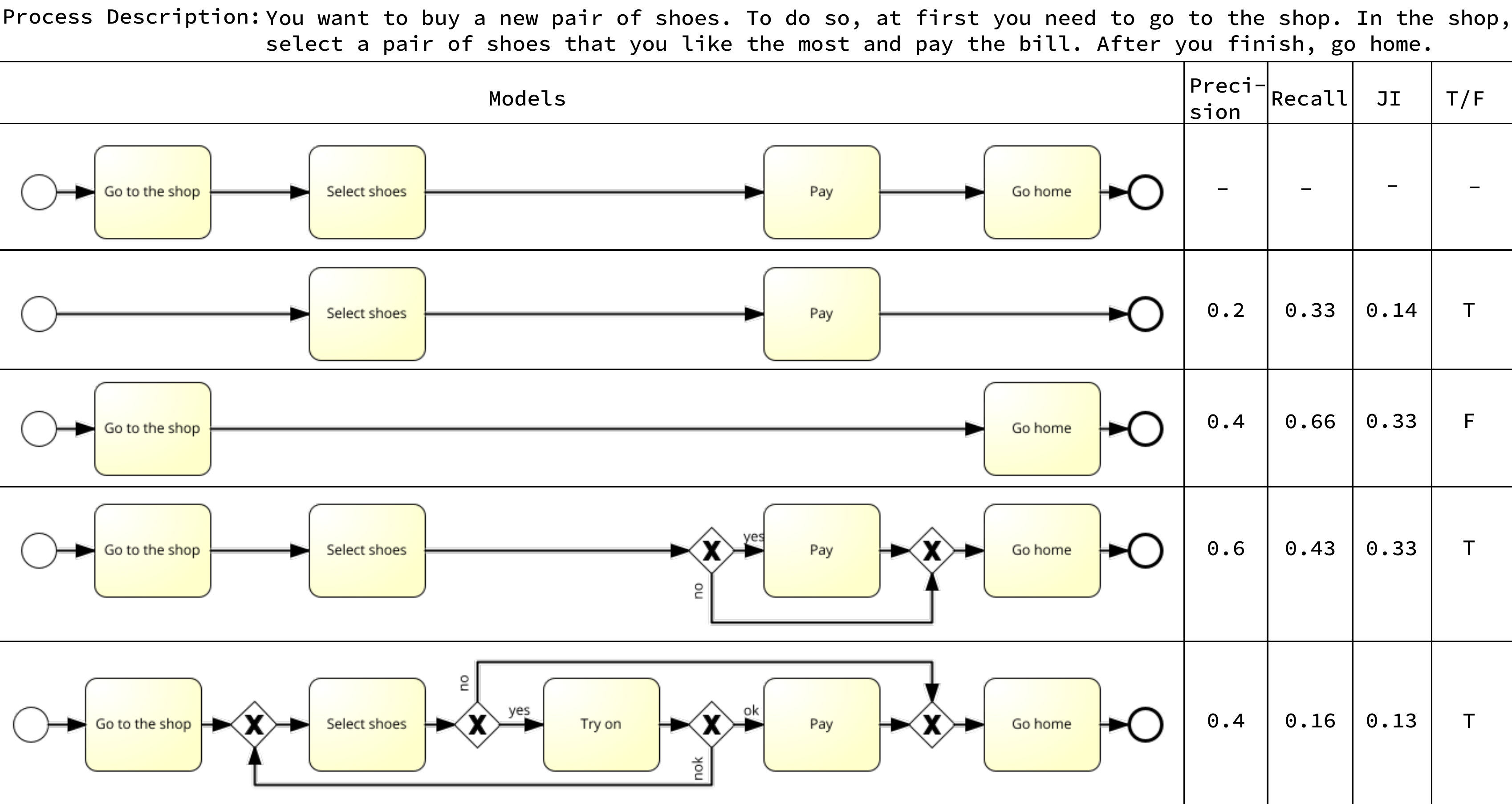}
    \caption{Model Variability Example: Different Granularities and Perspectives}
    \label{fig:variability}
\end{figure}

A process model is a bridge between a process and a customer involved in the process. Depending on a customer's tasks and goals, a modeller has to focus on the relevant aspects of a business process. It results in a model doubling (see Fig.~\ref{fig:variability}), where a single process can have multiple variations from a different perspective, level of abstraction and granularity~\citep{bpm_abstraction}. 

Taking this model variability phenomenon into account complicates model evaluation. For instance, if we consider the results presented in Tables~\ref{tab:jacc} -~\ref{tab:recall}, models designed by a human modeller have at least $10$\%\jvb{-points} lower metric values in comparison with the values gained by an automated system. The precision of a human modeller was two times worse than that of an automated system, reaching only 38\%. 

Low precision indicates that a human modeller tends to use another level of abstraction during process model design and tends to add further tasks, i.e., read between the lines. \jvb{Note that} the standard models are created based on human annotations directly derived from text. These process descriptions are fed into an automated system or a language model. Thus, quantitative metrics cannot be considered objective and universal (see Fig.~\ref{fig:variability}). Quantitative metrics could be used to evaluate the completeness of the model \jvb{with respect to the} original process description, but not its correctness.

\begin{table}[!ht]
    \centering
    \caption{Correctness of models generated by GPT3 and GPT4 based on process description}
    \begin{tabular}{l|l|l|l|l||l|l}
        \textbf{LLM} & \textbf{TR} &  \textbf{R} & \textbf{RA} & \textbf{RAB} & \textbf{F. Human} & \textbf{F. System} \\ \hline
        \multirow{2}{*}{GPT3} & GV & 2 & 3 & 2 & 7 & 4 \\ 
                              & MER & 3 & 2 & 1 & 7 & 4 \\ \hline
        \multirow{2}{*}{GPT4} & GV & 6 & 4 & 2 & 7 & 4 \\ 
                              & MER & 6 & 6 & 5 & 7 & 4 \\ 
    \end{tabular}
    \label{tab:cor}
\end{table}

Hence, every model is evaluated by a human to determine whether the models generated by LLMs are truly correct.
A `true' or `false' value is assigned to every model. A model is estimated to be correct if all logical aspects of the process description are captured. If some elements are missing, substituted, or new elements are added to the model and these changes do not violate the information introduced in the process description, the model is considered correct (see Fig.~\ref{fig:variability}). 

Table~\ref{tab:cor} shows the average results of models generated by GPT3 and GPT4 by executing (R),(RA) and (RAB) prompts. Each row represents the number of correctly generated models out of seven selected PET process descriptions for a particular prompt. 

We assume all models designed by human modellers are correct. None of the LLMs are able to achieve the same result as human modellers. GPT3 is able to generate only three correct models out of seven possible with both MER and GV representations, which is one model worse than automated systems. However, GPT4 generated six logically correct models. That is only one model less, when compared to a human modeller's result and $30$\%\jvb{-points} better as opposed to an automated system from the benchmark~\citep{friedrich} (see Sect.~\ref{sec:controlflow}). 

\begin{table}[!ht]
    \centering
    \caption{Average Ratio of Correctly Generated Models per Case}
    \begin{tabular}{l||l|l||l|l|l}
        Case & \# Words & \# Tasks & GPT3 & GPT4 & AVG \\ \hline
        10.13 & 39 & 3 & 0.33 & 0.67 & 0.50 \\ \hline
        10.6 & 30 & 4 & 0.67 & 1.00 & 0.83 \\ \hline
        10.1 & 29 & 4 & 0.67 & 1.00 & 0.83 \\ \hline
        5.2 & 83 & 7 & 0.50 & 0.67 & 0.58 \\ \hline
        3.3 & 71 & 7 & 0.00 & 0.83 & 0.42 \\ \hline
        1.3 & 162 & 11 & 0.00 & 0.00 & 0.00 \\ \hline
        1.2 & 100 & 10 & 0.00 & 0.67 & 0.33 \\
    \end{tabular}
    \label{tab:cases}
\end{table}

Regarding the correctness of models grouped by cases, the worst result occurred for case 1.3 in Table~\ref{tab:cases}. In this case, none of the generated models are correct. The underlying textual description is the longest and has the highest number of tasks. The best results are achieved by both GPT3 and GPT4 for cases 10.1 and 10.6. These descriptions are the shortest. These observations suggest a correlation between the complexity of the process and the quality of the generated model. However, not only the complexity of the process is critical. Case 10.13 is one of the shortest and most simple, containing three tasks. However, $50$\% of the models generated for this case are incorrect.

\nk{The possible explanation for this bad performance may be poor clarity, high complexity, and bad readability of the given process description for case 10.13. For instance, unnecessary repetitions, ambiguities, complex and unnatural sentence structures, unclear relationships among the steps (i.e., tasks) involved in the process, and excrescent information that is not expected in the model.} 
After paraphrasing process descriptions (reducing ambiguity and simplifying the structure of sentences) for both, case 1.3 and case 10.13, we generate new models with GPT4 using (R) and (RA) prompts. This lead to much better results, i.e., out of 4 models for 1.3, 2 models are correctly generated, while out of 4 models in 10.13, all 4 models are generated correctly. 

\nk{This implies that, above all, it is essential to convey the process using straightforward and unambiguous language to enhance the comprehensibility and readability of its description in order to achieve a high-quality model.}

\begin{table}[!ht]
    \centering
    \caption{Number of Correctly Generated Models (NCM) for all Cases}
    \begin{tabular}{l||l|l|l|l}
        LLM & \multicolumn{2}{c|}{GPT3} & \multicolumn{2}{c}{GPT4} \\ \hline
        TR & GV & MER & GV & MER \\ \hline
        NCM & 7 & 6 & 12 & 17 \\ \hline
        Accuracy in \% & 33 & 29 & 50 & 81 \\ 
    \end{tabular}
    \label{tab:accuracy}
\end{table}

\textbf{Summary.} Considering the maximum number of correctly generated models, both types of textual representations got the same results. Nonetheless, taking into account the number of correct generated models, GPT4 produces $17$ correct models out of $21$ with MER, reaching a performance of $81$\%. GV achieved a $57$\% accuracy. Using GPT3, the average accuracy of $30$\% is reached (see Table~\ref{tab:accuracy}).

\phantomsection
\addcontentsline{toc}{section}{another section in the document}
\label{corcomp}
\nk{Both the completeness and correctness of the model are vital characteristics, but they alone do not provide sufficient information. For instance, comparing correctly estimated models to incorrect ones, the difference in average C and NC Text Similarity is only 15 and 11\%-points, respectively (82\% compared to 67\%, and 42\% compared to 31\%, respectively). Model generated with LLM cannot be correct without being complete up to a certain level, and vice versa.}

\subsection{Evaluation: Survey}
\label{sub:survey}

\nk{In order to not rely only on quantitative evaluations and to exclude any prejudices and preconceptions during qualitative evaluation of models generated by LLMs we decided to conduct a survey\footnote{https://forms.office.com/e/Y55jyNuPi2} with students and professionals  with different programming and modelling background to select the process model (out of a list of given process models, that are considered as correct and complete, cf. Sect.~\hyperref[corcomp]{5.3}) that corresponds best to the provided process description. }

\nk{\srm{\textbf{Participant background:}} 29 out of 40 participants have more than 3 years of programming experience. 17.5\% of all participants are not familiar with markup languages. Out of those familiar with markup languages, 15\% have used MER, 42\% have used GV. Around 60\% of respondents are familiar with various graphical modelling languages (e.g., UML, ER, or BPMN). 80\% of them have more than 3 years of modelling experience or were applying modelling languages in class and industry projects and could be considered as confident modellers. Other participants either have no modelling experience or have few modelling skills.}

\phantomsection
\addcontentsline{toc}{subsection}{one more section}
\label{artifact}
\nk{\textbf{Artefact selection:} To be able to create a better process model, participants were asked to select one of the proposed artefacts or their combination (i.e., (r) a set of rules of how to represent a particular text as a graph; (a) an explicit list of process activities; (b) a summary of the BPMN standard) in addition to a textual process description. }

\nk{Most of the respondents select such artefacts as (a) or (r)+(a) (22.5\% each). Only 10\% of all participants suggest including additional artefacts or methods as process model examples or a workshop with a domain expert. Further, 12.5\% of all respondents suggest using other combinations of proposed artefacts like (a)+(b) or (r)+(b), although theses combinations are out of scope of this work. Best results were achieved by using either (R) or (RA) prompts, however only 7.5\% of all respondents consider using (r) independently.}

\nk{\textbf{Textual representation:} Respondents were also asked to rate their level of understanding when utilising (r) (i.e., MER and GV TNs) and (b) with respect to the 5-point scale in the survey. Out of the 40 participants, only 2 rated (b) as poor. Out of 26 respondents, that were rating (r), 23\% (MER) and 38\% (GV) consider them as unclear (i.e., rated as poor or very poor). Nevertheless, 70\% of respondents rate MER (r) as good or very good. At the same time, only 30\% of participants acknowledge GV (r) as good or very good.}

\nk{From 26 participants who evaluated MER and GV (r), only 3 and 9 respondents respectively, utilised textual representations. These respondents are distinct individuals, with no overlap between the two groups. Interestingly, a significant majority, 65\% (17 participants), find MER to be easier and expressed a preference for using it over GV. In contrast, only 20\% (5 participants) prefer GV, and 15\% (4 participants) did not have any preference.}

\nk{However, it is essential to highlight that the MER/GV preference distribution might be linked to the prior experience and comprehension of the provided (r) artefacts, observed during artefact selection (see Sect.~\hyperref[artifact]{5.4}).}

Please note that in a highly integrated prototype for using LLMs to create graphical models out of text, neither GV nor MER (r) would be visible to the user. \nk{Even so, one thing to highlight is that LLMs demonstrate a reduced occurrence of syntax errors using GV's textual representation.}

\nk{\textbf{Hypothesis testing:} To assess whether LLM generated models could be recognised as the models, designed by a human modeller we ask participants to explore a process description and select one of proposed models, that in their opinion corresponds best to it. Only 5 out of 7 cases were picked, as for cases 10\_1 and 10\_6 all correctly created models were identical with the original ones. For every question participants could select one model out of between 2 and 5 models, where one model is always the original one and the others are generated by LLMs.}

\nk{To determine whether the actual data conforms to an expected pattern or distribution the Pearson's chi-squared goodness-of-fit test is used. This test is chosen because we are examining just one categorical variable and comparing its observed frequencies to expected frequencies. Mostly, sources suggest to apply the chi-squared test for larger sample sizes, but there's no agreement on a 'large' and 'small' sample size definition or its boundaries~\citep{chi}. According to~\citep{chirec} no expected frequency should fall below 5.}

\nk{For every case, all collected responses ($N=40$) fall into three groups: standard model (stand), LLM generated model (llm), or other (i.e., a model suggested by a user that was not included in the proposed options). The ``I don't know'' option was removed from the observations since, in the majority of cases, participants tend to select it for all the cases (or its majority) due to limited experience, lack of motivation, or time constraints. Only a small number of all responses (10\%) correspond to the ``I don't know'' option.}

\nk{Our null hypothesis is based on the assumption that the majority of people prefer the standard model over LLM generated and states the probabilities for each group as follows:
\begin{equation}
H_0: P = (P_1,P_2,P_3) = (0.8,0.15,0.05) 
\end{equation}
, where $P_1$ is the probability that the standard model is selected, $P_2$ stands for the LLM-generated model, and $P_3$ for ``other''.}

\nk{Our alternative hypothesis suggest that the distribution of responses differs from our null hypothesis. In other words, we suspect that people do not predominantly select the standard model, and there is a change in the distribution among the three groups:
\begin{equation}
H_1: P \neq (0.8,0.15,0.05)
\end{equation}}

\begin{figure}[ht]
    \centering
    \includegraphics[width=0.75\textwidth]{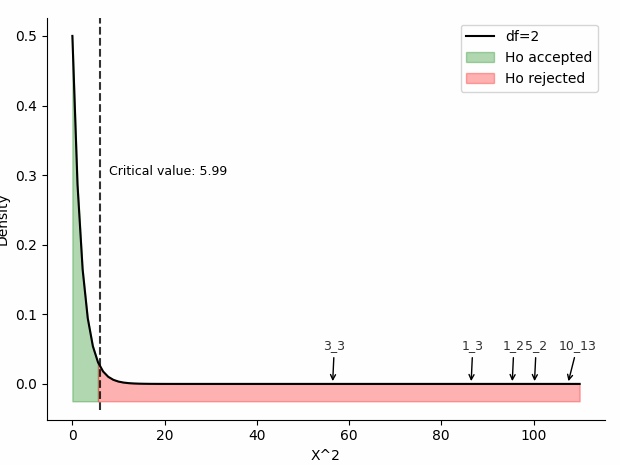}
    \caption{Pearson’s Chi-Squared Test Results: Model Preference}
    \label{fig:chi}
\end{figure}

\nk{To determine the validity of the null hypothesis and whether it should be rejected, we compute the critical value for the goodness-of-fit test and compare it to the value obtained for each case. If the goodness-of-fit value for a case is greater than  the critical value (i.e., falls into the rejection region) and the p-value is less than selected significance level (= 0.05), then we should reject the null hypothesis (see Fig.~\ref{fig:chi}).}

\nk{In every case, the goodness-of-fit is significantly larger as its critical value of 5.99, and the p-value is well below the selected level of significance (p=0.0001). These findings could be considered as highly significant, indicating that the null hypotheses should be rejected.}

\nk{It appears that only 19\% of all responses correctly recognise the standard model, while in 69\% of all responses models generated by LLMs were selected as a correct ones.}

\nk{Since only half of the participants possess modelling experience and are familiar with BPMN, we aim to investigate whether an association exists between the choice of model type (LLM, standard, or other) and the level of programming experience (i.e., (ne) no experience, (crb) learned in class or from book, (cp) used on a class project, (pi) used on one project in industry, (mp) used on multiple projects in industry). Our inference is that individuals with greater modelling experience are more likely to choose ``other'' or ``standard'' choices, while those with less experience tend to prefer LLM-generated models.}

\nk{To investigate whether there is a true relation between modelling experience and selected model type, we employ two tests of independence: the Pearson's chi-squared test and the Fisher's exact test. In both cases, our null hypothesis states that the variables are independent from each other:
\begin{equation}
H_0: A \perp B
\end{equation}, where A is type of chosen model and B is the level of modelling experience.}

Conversely, the alternative hypothesis claims that there is a connection between A and B.

Considering case individual results for both Fisher and Pearson tests (see Table~\ref{tab:independent}) in 9 out of 10 cases the null hypothesis is accepted. This indicates that we don't have enough confirmation to claim that there is a connection between the chosen model type and a level of modelling skills. In simpler terms, these two variables are considered to be independent.

\begin{table}[!ht]
    \centering
    \caption{Relationship Between Modelling Experience and Selected Model Type: p-values}
    \begin{tabular}{l|l|l}
        Case & Fisher & Pearson \\ \hline
        1\_2 & 0.387 & 0.306 \\ \hline
        1\_3 & 0.129 & 0.179 \\ \hline
        3\_3 & 0.030 & 0.053 \\ \hline
        5\_2 & 0.336 & 0.292 \\ \hline
        10\_13 & 0.06 & 0.111 \\ \hline
         \hline
        total & 0.044 & 0.054 \\ 
    \end{tabular}
    \label{tab:independent}
\end{table}

\nk{However, when we consider all cases collectively as a single sample, the overall picture undergoes a reasonable shift. According to the Fisher test, the null hypothesis is expected to be rejected, suggesting a potential dependency between modelling experience and choice of a model type. Based on the Pearson test, the null hypothesis remains valid, but the p-value is only slightly higher than the initially set level of significance.}

\begin{figure}[ht]
    \centering
    \includegraphics[width=0.7\textwidth]{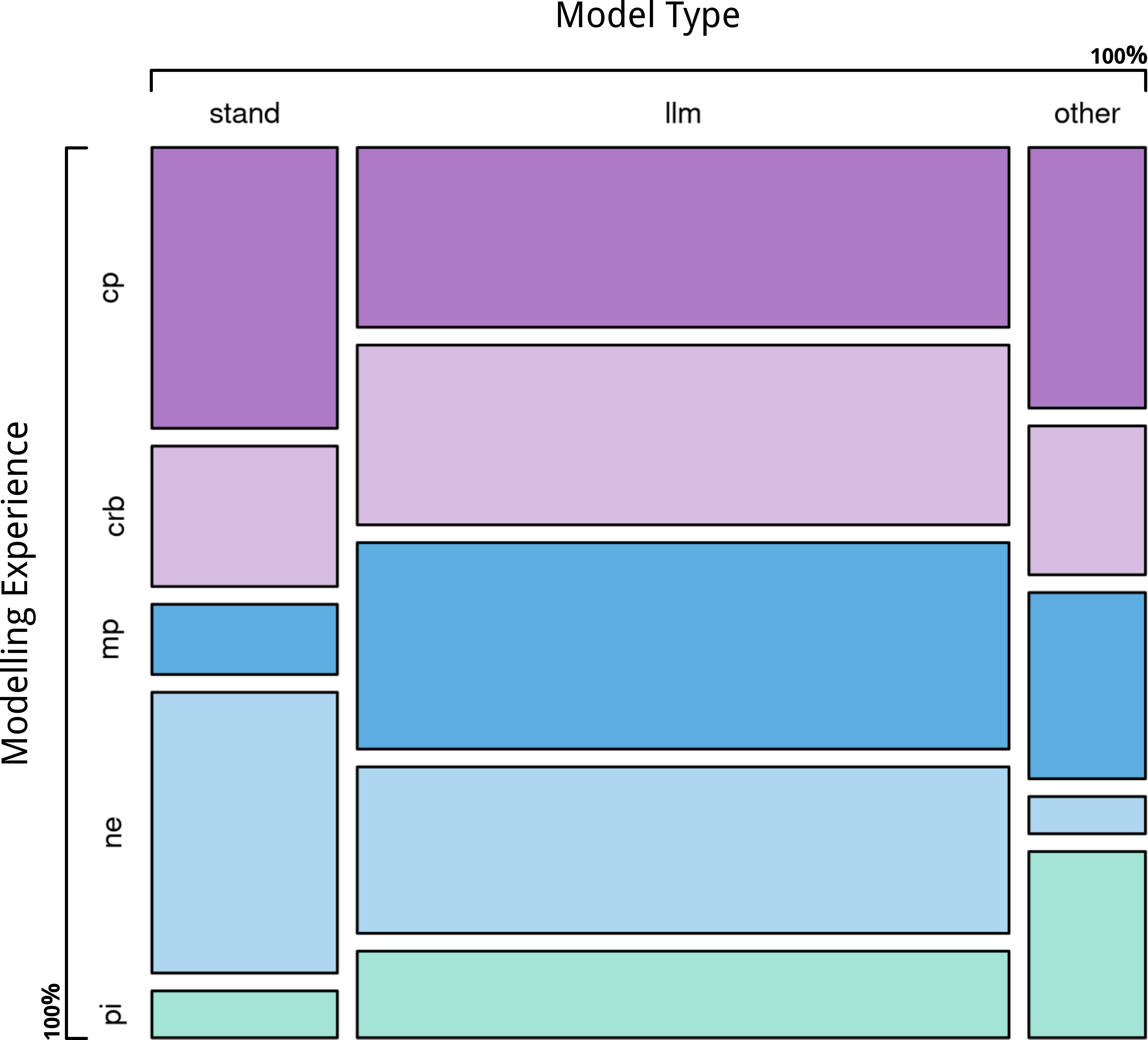}
    \caption{Percentage Distribution of Frequencies For Modelling Experience and Selected Model Type}
    \label{fig:chi-mos}
\end{figure}

\nk{An intriguing finding is that the current relationship between model type and modeling experience differs from our initial expectations. Participants with no modeling experience (ne) and those with a more academic background (crb and cp) tend to choose standard models more frequently compared to individuals with real-life experience (pi and mp). Conversely, more experienced respondents show a preference for the 'other' option over the standard models. However, all groups consistently vote for LLM-generated models. The distribution of LLM model selections remains consistent (~60\%) across all five groups (see Fig.~\ref{fig:chi-mos} for more details).}

\nk{\textbf{Concerns:} It is essential to mention that these results should be considered with caution, due to relatively small sample size. With half of the respondents being not or limited familiar with the subject matter, it is important to acknowledge the potential influence of various response biases. Survey respondents' engagement and data quality can be compromised by survey length and complexity, potentially resulting in rushed or incomplete responses. Misinterpretation of questions and the influence of social desirability bias may introduce inaccuracies. }

\subsection{Control Evaluation}
\label{sec:control}
Since all prompts and their corresponding artefacts are generated using the PET dataset and a trial-and-error approach, we decide to conduct a control evaluation with our initial dataset (see Sect.~\ref{sec:texset}). This is done to check whether considered prompts and artefacts can yield the optimal results with other data, or they are affected by potential biases, overfitting, or lack of generalisation. In addition, due to our limited knowledge about the training process of GPT models, it is impossible to determine whether the LLM has been trained with any particular dataset already~\citep{reason}.

\nk{For this purpose only the best results from quantitative and qualitative evaluations, as well as survey feedback, are considered, i.e., GPT4 with MER, (R), and (RA) prompts. Every generated model is labelled as correct if it adheres to the BPMN standard and does not include any elements or relations, that contradict the original process description. In addition to correctness, we take completeness into account (see Sect.~\ref{sec:taskextraction}).}

\nk{Despite the fact that with both (R) and (RA) the average text similarity of 74\% was reached, utilising (R) prompt, 10 models out of 21 are generated correctly, and with (RA) only 8 (see~\footref{arepo}). The average achieved accuracy is only 43\%, which is half of what was attained using the PET dataset (see Table~\ref{tab:accuracy}).}

\nk{Surprisingly, only 52\% of all models got the same labels for the same process description, i.e., they are either both considered as correct or incorrect. The 50\% variation in labelling can be attributed to the ambiguity of language, the complexity of interpreting tasks, and the model's inherent variability in processing such tasks.}

\nk{Basically, all mistakes can be categorised into three main groups: 
\begin{enumerate}
\item[(1)] \textit{violations of provided rules and standards}: the absence of BPMN elements such as gateways, start, or end events where they are required, end events placed in the middle of the process, or the use of splitting and merging gateways of different types; 
\item[(2)] \textit{mistakes occurred from incomplete information in the process description}: gateways with a single path or abrupt branches without closure, treating examples within the process description as separate tasks and integrating them into the control flow, resulting in inconsistencies; 
\item[(3)] \textit{discrepancies in the control flow unrelated to the textual description}: deviations in task sequence, misplacement's of decisions, etc. 
\end{enumerate}
}

\begin{figure}[ht]
    \centering
    \includegraphics[width=0.7\textwidth]{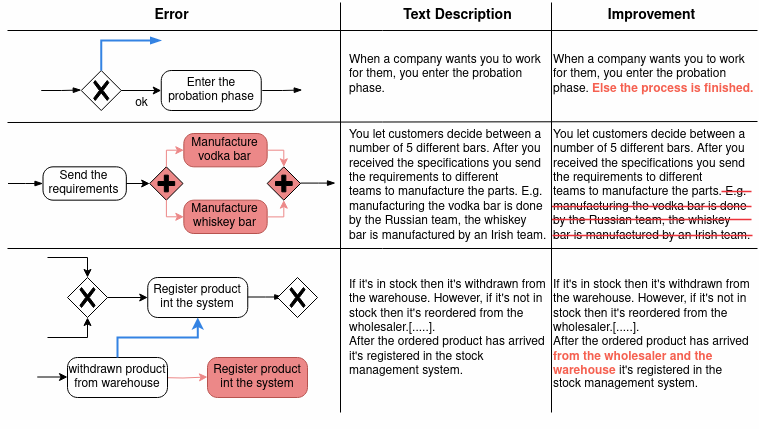}
    \caption{Common Errors (Group 2) Occurring during Model Generation by LLM: Red Denotes Wrongly Generated Elements, Blue Denotes Missing Elements.}
    \label{fig:bpmn-mis}
\end{figure}

\nk{Groups (1) and (2) can be typically corrected/avoided (see Fir.~\ref{fig:bpmn-mis}) by improving the structure and clarity of the provided artefacts, such as textual process descriptions, rules for the selected textual representation of a model, and descriptions of standards. Mistakes falling into group (3) are often unpredictable and usually require human interaction for correction.}

\subsection{Discussion}
\label{sec:disc}

A first insight is that manually designed and refined models contain additional tasks that cannot be directly extracted from the original text, but exist due to a humans' ability to ``read between the lines'' or reason about task granularity. GPTs tend to extract tasks at the level of abstraction as provided by the input text and do not have the capability to reason about when it makes more sense to have multiple small tasks instead of one big one. We tried to coax GPTs into extracting more tasks by restricting the number of words describing a task (i.e., its label), which increased the average number of extracted tasks slightly by 1 for GPT1-3.5, and by 2 for GPT4, as can be seen in Table~\ref{tab:restsetsim}.

On average, GPTs extract a third less tasks than existed in the process models. When strictly looking at the capability of extracting tasks from the original text, GPT3 and GPT4, on average, achieves a text similarity of $80$\% and 85\%, respectively. The interpretation of this value is difficult. It could mean that the LLM missed about $15-20$\% of the tasks or, alternatively, that $15-20$\% of the text are just the filler words that have been ignored by the LLM. Together with the observation that the LLM tends to not split up tasks, the $30$\% less tasks extracted from the text in comparison to the model, hint at a possible explanation.

Considering only the completeness of a model (quantitative metrics from Sect.~\ref{sec:controlflow}) there \jvb{is no substantial} difference between the performance of GPT3 and GPT4. However, looking at both completeness and correctness of a model (qualitative and quantitative metrics) GPT4 is significantly better ($50$\%) than GPT3. With MER, more correct models are generated in comparison to GV. Despite the fact, that LLMs seem to be more confident reaching better scores with MER representations (see Sect.~\ref{sub:correctness},~\ref{sec:eval}), over $70$\% of all the models created with MER contain syntactical errors or additional content. At the same time, GV representations are always generated correctly. The reason for such a behaviour can be either poor lucidity and clarity of provided rules for this particular textual representation, or it could indicate, that LLMs are more familiar with Graphviz syntax, as Graphviz is older and may hence be more prevalent in training data.

\nk{Not only factors such as selected textual representation, complexity of the provided process and instructions, the length of textual description have an influence on the quality of a generated model, but also the clarity and the structure of the process description itself are important. The existing dependency between the quality of a generated model and its textual description (see Sect.~\ref{sub:correctness}) emphasises that LLM models generate human-like text based on predefined patterns but may not possess true comprehension (i.e., can LLM models really 'reason' or it is still just a stochastic parrot) and leads to a reasonable question, whether LLMs can be used for specific tasks, where human understanding and experience are required. }

\nk{On the other hand, according to the survey results, almost 70\% of participant prefer models, generated by LLMs over two other options: either model created based on human annotations or suggesting their solutions (see Sect.~\ref{sub:survey}). This preference remains consistent regardless modelling experience of all the respondents. }

\nk{Nevertheless, the reliability of GPT models in consistent producing complete and correct models is questionable, as the accuracy of correctly generated models declines to an average of 43\% during control evaluation (see Sect.~\ref{sec:control}).}

Finally, let us highlight that the empirical results merely represent a snapshot in time of (largely) commercial LLM capabilities.
In order to assess the frontier of what is currently possible, we opted to focus our assessment on proprietary, closed LLMs (or rather: on the commercial product capabilities containing these LLMs) instead of recent, yet currently arguably weaker, openly available LLMs\footnote{https://huggingface.co/spaces/lmsys/chatbot-arena-leaderboard}. Still, let us highlight that we not only provide an assessment of capabilities, but also a general methodological framework for conducting assessments of ConverMod capabilities, thus laying the foundation for future research on both proprietary and open models.

\section{Conclusion: Practical Implications and Research Directions}
\label{ref:vision}

From the state-of-the-art discussion in Sect. \ref{sec:sota} and the results of the evaluation presented in Sect. \ref{sec:disc}, the following two main managerial implications can be derived:

\begin{enumerate}
    \item For the chatbot application scenarios ``gather information'' and ``process modeling'' (cf. Table \ref{tab:applife}), chatbots are in principle ready to be applied in practice as-is, yet the results have to be taken with a grain of salt, i.e., the domain expert should always check the results. \jvb{However, the checking of the generated model by the domain expert can still lead to variations in model quality, as the survey in Sect. \ref{sub:survey} indicates that a standard, ground truth model is not easily selected from a set of generated models. Interestingly, the selection ability of humans does not necessarily depend on the modelling experience. Additionally, the particular chatbot implementation should take into account the complex relationships between model completeness, correctness, \nk{textual process description}, textual representation and prompt engineering such that the modelling capability of the chatbot can be tuned towards the respective needs of the domain expert (cf. Sect. \ref{sub:correctness}). The results for these relationships given the quantitative and qualitative evaluation in Sect. \ref{sec:eval} and Sect. \ref{sub:correctness} should either be risen to the awareness of the domain expert via training before roll-out of the chatbot or added to the conversational user interface of the chatbot, e.g., by means of selecting the textual representation of the generated model.}
    \item For the chatbot application scenarios ``compare and assess'', ``select method, query models'', and ``query and refactor models'', off-the-shelf chatbots are not yet ready to be applied due to their inability to understand process model semantics. \jvb{Still, a combination of chatbot based model generation with syntax checking/reasoning capabilities of existing tools such as SAP Signavio or ProM can be a way to overcome the current barriers to apply chatbots in these application scenarios.}
\end{enumerate}

As business process modelling has become an important tool for managing organisational change and for capturing requirements of software, the first managerial implication is that ConverMod can already have a significant business impact. Considering that the central problem in this area -- the acquisition of as-is models -- consumes up to 60\% of the time spent on process management projects~\citep{bpmngen}, chatbot-based partial automation can be sufficiently impactful, even if substantial human refinement is required.

The second managerial implication is that future research should focus on integrating the strong language capabilities of chatbots with the specialised capabilities of existing knowledge-based tools. The integrative research direction is more promising than chatbot training with specialised process modelling training sets featuring native process models, e.g., process models in BPMN format and a number of semantic targets, such as information on the existence of deadlocks in a process model. First, training of the chatbot with respect to business process models ignores the vast existing modelling knowledge encoded into existing tools. Second, semantics are clearly defined and encoded in existing tools such that training chatbots with the aim of understanding formal semantics is futile unless it serves as an intermediate step that unlocks further value.

To conclude, while advanced tasks such as model querying, refinement, and analysis presumably require domain-specific solutions, the interplay of traditional, knowledge based approaches to business process modelling can relatively straight-forwardly be augmented by machine learning-based chatbots to facilitate tedious tasks such as information gathering and basic model creation.

\end{document}